\documentclass[10pt,journal,final,compsoc]{IEEEtran}
\ifCLASSOPTIONcompsoc
  \usepackage[nocompress]{cite}
\else
  \usepackage[noadjust]{cite}
\fi
\ifCLASSINFOpdf
\else
\fi
\usepackage{multirow}

\usepackage[table]{xcolor}

\usepackage{graphicx} 
\usepackage{ulem}

\usepackage{subfig}
\usepackage{ragged2e}
\renewcommand{\raggedright}{\leftskip=0pt \rightskip=0pt plus 0cm}

\usepackage{amssymb}
\usepackage{amsthm}

\newtheorem*{remark}{Remark}
\usepackage[tbtags]{amsmath}
\newtheorem{definition}{Definition}

\DeclareMathOperator*{\argmin}{arg\,min}
\DeclareMathOperator*{\argmax}{arg\,max}

\usepackage{mathtools}
\DeclarePairedDelimiterX{\infdivx}[2]{(}{)}{%
  #1\;\delimsize\|\;#2%
}

\DeclarePairedDelimiter{\norm}{\lVert}{\rVert}
\newcommand{\Tau}{\mathcal{T}}

\usepackage{algorithm}
\usepackage{algpseudocode}
\algnewcommand{\Initialize}[1]{%
  \State \textbf{Initialize:}
  \Statex \hspace*{\algorithmicindent}\parbox[t]{.8\linewidth}{\raggedright #1}
}
\algnewcommand{\Inputs}[1]{%
  \State \textbf{Inputs:}
  \Statex \hspace*{\algorithmicindent}\parbox[t]{.8\linewidth}{\raggedright #1}
}
\algnewcommand{\Outputs}[1]{%
  \State \textbf{Outputs:}
  \Statex \hspace*{\algorithmicindent}\parbox[t]{.8\linewidth}{\raggedright #1}
}
\newtheorem{theorem}{Theorem}[section]

\newcommand{\eat}[1]{}

\usepackage{booktabs,siunitx}

\usepackage{amsthm}
\usepackage{bbm}
\usepackage{mathrsfs}
	
\usepackage{color, colortbl}
	
\definecolor{LightCyan}{rgb}{0.88,1,1}

\DeclareMathOperator{\rank}{rank}
\usepackage[colorlinks]{hyperref}
\usepackage[colorinlistoftodos]{todonotes}
\usepackage{url}

\begin{document}
%
\title{Source-Free Progressive Graph Learning for Open-Set Domain Adaptation}
%
%
%
%

\author{Yadan~Luo, Zijian~Wang, Zhuoxiao~Chen, Zi~Huang,
        and Mahsa~Baktashmotlagh
\IEEEcompsocitemizethanks{\IEEEcompsocthanksitem Y. Luo, Z. Wang, Z. Chen,  Z. Huang and M. Baktashmotlagh were with School of Information Technology and Electrical Engineering, The University of Queensland, Australia\protect\\
E-mail: \{y.luo, zijian.wang, zhuoxiao.chen, helen.huang\}@uq.edu.au, m.baktashmotlagh@uq.edu.au.}
\thanks{Manuscript received April 19, 2005;}}

%
%

\markboth{IEEE TRANSACTIONS ON PATTERN ANALYSIS AND MACHINE INTELLIGENCE}%
{Shell \MakeLowercase{\textit{et al.}}: Bare Demo of IEEEtran.cls for Computer Society Journals}
%




\IEEEtitleabstractindextext{%
\begin{abstract}

\raggedright Open-set domain adaptation (OSDA) has gained considerable attention in many visual recognition tasks. The aim of OSDA is to transfer knowledge from a label-rich source domain to a label-scarce target domain while addressing the disturbances from the irrelevant target classes that are not present in the source data. However, most existing OSDA approaches are limited due to three main reasons, including: (1) the lack of essential theoretical analysis of generalization bound, (2) the reliance on the coexistence of source and target data during adaptation, and (3) failing to accurately estimate the uncertainty of model predictions. To address the aforementioned issues, we propose a Progressive Graph Learning (PGL) framework that decomposes the target hypothesis space into the shared and unknown subspaces, and then progressively pseudo-labels the most confident known samples from the target domain for hypothesis adaptation. The proposed framework guarantees a tight upper bound of the target error by integrating a graph neural network with episodic training to suppress underlying conditional shift, as well as leveraging adversarial learning to close the gap between the source and target distributions. Moreover, we tackle a more realistic source-free open-set domain adaptation (SF-OSDA) setting that makes no assumption about the coexistence of source and target domains, and introduce a balanced pseudo-labeling (BP-L) strategy in a two-stage framework, namely SF-PGL. Different from PGL that applies a class-agnostic constant threshold for all target samples for pseudo-labeling, the SF-PGL model uniformly selects the most confident target instances from each category at a fixed ratio. The confidence thresholds in each class are regarded as the `uncertainty' of learning the semantic information, which are then used to weigh the classification loss in the adaptation step. We conducted unsupervised and semi-supervised OSDA and SF-OSDA experiments on the benchmark image classification and action recognition datasets. The reported results evidence the superiority and flexibility of the proposed PGL and SF-PGL methods in recognizing both shared and unknown categories. Additionally, we find that balanced pseudo-labeling plays a significant role in improving calibration, which makes the trained model less prone to over-confident or under-confident predictions on the target data. Source code is available at \url{https://github.com/Luoyadan/SF-PGL}.
\end{abstract}

}

\maketitle

\IEEEdisplaynontitleabstractindextext

%
\IEEEpeerreviewmaketitle

\IEEEraisesectionheading{\section{Introduction}\label{sec:introduction}}

%
%
%
%
\IEEEPARstart{W}{hile} 
deep learning has made remarkable advances across a wide variety of machine-learning tasks and applications such as image and video recognition, it is commonly at a great cost of curating large-scale training data annotations. To relieve the burden of expensive data labeling, transfer learning has been introduced to extract knowledge from the existing annotated training data (i.e. source domain) and convey it to the unlabeled or partially labeled test data (i.e. target domain). However, the source and target domains are generally constructed under varying conditions such as illuminations, camera poses, and backgrounds, which is referred to as \textit{domain shift}. For instance, the Gameplay-Kinetics~\cite{DBLP:conf/iccv/ChenKAYCZ19} dataset for action recognition is built under the challenging ``Synthetic-to-Real'' protocol, where the training videos are synthesized by game engines and the test samples are collected from real scenes. In this case, the domain shift between the source and target domains inevitably leads to severe degradation of the model generalization performance. 

To mitigate the aforementioned domain gap, unsupervised domain adaptation (UDA) techniques have been proposed to align source and target distributions through statistical matching~\cite{dip, manifold, distribution, DDC, JDA, DAN, DANN,ADDA,jingjing} or adversarial learning~\cite{DANN,ADDA,JAN,DRCN,DBLP:conf/mm/LuoHW0B20,DBLP:conf/mm/WangLHB20}, which provide rigorous error bounds on the target data~\cite{david1, discrepancy, bridgingtheory}. Although the UDA methods have been advanced and applied on many tasks such as object detection, semantic segmentation, and action recognition, the evaluation protocols were restricted to a scenario where the target domain shares an identical set of classes with the source domain. Such a scenario typically refers to a \textit{closed-set} setting, which could be hardly guaranteed in real-world applications, where the test samples may come from unknown classes that are not seen during training.

In the light of the above discussion, a more realistic \textit{open-set} domain adaptation (OSDA) setting \cite{OSBP} has been introduced, which allows the target data to contain an additional ``unknown'' category, covering all classes that are not present in the source domain. The key challenge of OSDA is to safely transfer knowledge across the domains while recognizing the unknown classes accurately. To tackle this, different strategies such as confidence manipulation \cite{OSBP,Attract-UTS}, subspace reconstruction \cite{mahsa}, instance weighting \cite{STA}, and extreme value theory \cite{DBLP:conf/aaai/Jing0ZDLY21} have been studied. Nevertheless, there are three non-negligible obstacles that prevent existing OSDA methods to be applied successfully in a real-world scenario:

\begin{itemize}
\item The lack of theoretical analysis of generalization bound for OSDA methods: According to~\cite{david1, discrepancy}, the target error is bounded by four factors including the source risk, discrepancy across the domains, the shared error coming from the conditional shift~\cite{conditionalshift}, and the open-set risk. Among all, open-set risk contributes the most to the error bound, specifically when a large percentage of data belongs to the unknown class. However, designing an effective strategy to minimize the open-set risk remains an open problem.

\item The reliance on co-existing source and target data during adaptation: Deploying the existing OSDA approaches on portable and mobile devices is infeasible, as they require loading and processing large-scale source data. Source videos in the Gameplay~\cite{DBLP:conf/iccv/ChenKAYCZ19} dataset may consume hundreds of gigabytes of storage. In addition, the assumption of data accessibility is likely to trigger concerns for data sharing and digital privacy, especially in the medical and biometrics communities.

\item The failure of estimating predictive uncertainty of models: Solely focusing on improving target accuracy at inference time could result in producing over-confident predictions in mainstream OSDA methods. This issue, typically referred to as \textcolor{black}{a miscalibration  \cite{DBLP:conf/icml/GuoPSW17,DBLP:conf/aistats/ParkBWL20,DBLP:conf/nips/WangL0J20}}, can be a major problem in decision-critical scenarios.
\end{itemize}

In this paper, we, therefore, propose a generic Progressive Graph Learning (PGL) framework and its source-free variant (SF-PGL) for OSDA. We theoretically analyze the generalization and calibration properties of the proposed frameworks. The PGL method follows a common open-set domain adaptation setting, where the source and target data are available during training. The deep model consists of a feature extraction module, a graph neural network, and a classifier (hypothesis). To minimize partial risks and achieve a tighter error bound for open-set adaptation, the proposed PGL integrates four different strategies: (1) To suppress the source risk, we decompose the original hypothesis space $\mathcal{H}$ into two subspaces $\mathcal{H}_1$ and $\mathcal{H}_2$, where $\mathcal{H}_1$ includes classifiers for the shared classes of the source and target domains and $\mathcal{H}_2$ is specific to classifying unknowns in the target domain. With a restricted size of the subspace $\mathcal{H}_1$, the possibility of misclassifying source data as unknowns will be reduced. (2) To control the open-set risk, the progressive learning paradigm~\cite{curriculum} is adopted, where the target samples with low classification confidence are gradually rejected from the target domain and inserted as the pseudo-labeled unknown set in the source domain. This mechanism suppresses the potential negative transfer where the private representations across domains are falsely aligned. (3) To address conditional shift~\cite{conditionalshift} at both sample- and manifold-level, we design an episodic training scheme and align conditional distributions across domains by gradually replacing the source data with the pseudo-labeled known data in each episode. We learn class-specific representations by aggregating the source and target features and passing episodes through deep graph neural networks. (4) An adversarial domain discriminator is seamlessly equipped, which effectively closes the gap between the source and target marginal distributions for the known categories.

While effective, the PGL model relies on the adversarial learning mechanism to align cross-domain distributions, hereby failing to handle the source-free scenarios with no access to the source data. To overcome this limitation, we further put forward a \textit{balanced pseudo-labeling} (BP-L) strategy and form a complete two-stage framework, namely, SF-PGL. More specifically, we pass all target data to obtain its respective predictions through the freezed feature extraction module and the hypothesis, which are pre-trained by the source data. Then we sort the confidence of target predictions and evenly select a fixed-size group of high-confidence samples from each category to initialize the pseudo-labeled known set. The confidence threshold in each class is recorded, which measures the `uncertainty' of learning the class-specific information. In the second stage, the based model along with the graph neural network is iteratively trained with the pseudo-labeled and unlabeled target samples until all target samples are labeled. The predictions produced from the trained model are uncertainty-aware, due to the fact that the classification loss for each instance is weighted by the class uncertainty. 

A preliminary version of this work was presented in \cite{pgl}. In this work, we additionally (1) introduce a novel variant SF-PGL framework tailored for the source-free setting and provide a thorough experimental evaluation on its calibration capacity. The proposed SF-PGL model achieves the lowest expected calibration error (ECE) compared to all conventional open-set and source-free open-set domain adaptation methods. (2) We further apply the proposed PGL approach to four action recognition datasets i.e.,  \textit{UCF-HMDB$_{small}$}, \textit{UCF-Olympic}, \textit{UCF-HMDB$_{full}$} and \textit{Kinetics-Gameplay}, which are manually formed for the tasks of OSVDA and S-OSVDA. Our approaches achieve state-of-the-art results in all settings, both for unsupervised and source-free open set domain adaptation.






\section{Related Work}
\subsection{Open-set Domain Adaptation}
\noindent Different from canonical closed-set domain adaptation, open-set domain adaptation (OSDA) \cite{ATI,OSBP,DBLP:journals/pami/BustoIG20} addresses the interference from the unshared categories in the target domain when adapting the learned models. To avoid the potential risk of \textit{negative transfer} \cite{DBLP:journals/tkde/PanY10} brought by the unshared categories, it is important for OSDA methods to accurately determine the irrelevant target samples as the `unknown' class while aligning the shared classes. The most intuitive way is leveraging OSVM \cite{OSVM} that uses a class-wise confidence threshold to classify target instances into the shared classes, or reject them as unknown. Busto and Gall \cite{ATI} introduced an ATI-$\lambda$ method, which assigns the target data a pseudo class label or an unknown label based on its distance to each source cluster. Saito \textit{et al}. \cite{OSBP} derived an objective in the adversarial Open Set Back-Propagation (OSBP) framework, which balances the classifier's confidence on the known and unknown class with a threshold. Baktashmotlagh \textit{et al}. \cite{mahsa} proposed to learn factorized representations of the source and target data, so that unknown points can be identified by examining reconstructions from domain-specific subspaces. Liu \textit{et al}. \cite{STA} and Feng \textit{et al}. \cite{Attract-UTS} aimed to push the unknown class away from the decision boundary by a multi-binary classifier or semantic contrastive mapping. Luo \textit{et al}. \cite{pgl} followed the progressive learning paradigm, which globally ranks all target samples and gradually isolates the ones of lower confidence as unknown samples. Bucci \textit{et al}. \cite{DBLP:conf/eccv/BucciLT20} built the framework upon the recent success of self-supervised learning, which separates the unknowns with the confidence of predicting rotations and semantics. Jing \textit{et al}. \cite{DBLP:conf/aaai/Jing0ZDLY21} leveraged Distance-Rectified Weibull model for rejecting known samples based on the angular distance. Chen \textit{et al}. \cite{DBLP:conf/mmasia/ChenLB21} applied a class-conditional extreme value theory for open-set video domain adaptation. Of late, Universal Domain Adaptation (UDA) \cite{DBLP:conf/cvpr/YouLCWJ19} has been proposed, which further handles the case when the source domain holds private classes. While effective, all these methods assume the target user's access to the source domain, which could be infeasible due to privacy and security issues.

\subsection{Source-free Domain Adaptation}
Source-free domain adaptation studies how to transfer knowledge when the source data is absent. Kundu \textit{et al}. \cite{DBLP:conf/cvpr/KunduVRVB20,DBLP:conf/cvpr/KunduVVB20} proposed to generate out-of-domain (OOD) samples by applying the feature splicing technique, which enhances the model generalization capacity to the unseen samples in the target domain. Liang \textit{et al}. \cite{DBLP:conf/icml/LiangHF20} aimed to force the target representations to resemble the source features by information maximization while augmenting target features with self-supervised pseudo-labeling. Li \textit{et al}. proposed a 3C-GAN to augment the target sets for training, where the weight regularization and clustering based regularization are adopted for preventing classifier drifting and smoothing distribution. In the same vein, Kurmi \textit{et al}. \cite{DBLP:conf/wacv/KurmiSN21} leveraged generated data points as proxy samples for adaptation, where the generative model is modeled as an energy-based function. \textcolor{black}{The proposed approach differs from prior works by designing a balanced pseudo labelling which takes the class uncertainty into consideration. The pseudo labeling is progressively learned which alleviates accumulated errors and help the classifier well-calibrated. }
\textcolor{black}{
\subsection{Episodic Training}
The notion of `episodic training' \cite{DBLP:conf/iccv/LiZYLSH19,DBLP:conf/iccv/Qiao000HW19,DBLP:conf/nips/VinyalsBLKW16} originates from meta-learning \cite{DBLP:journals/air/VilaltaD02, DBLP:journals/pami/HospedalesAMS22, DBLP:conf/aaai/LuoHZWBY20, DBLP:conf/icml/FinnRKL19, DBLP:conf/icml/AcarZS21, DBLP:conf/icml/FinnAL17, DBLP:conf/nips/FinnXL18, DBLP:conf/iclr/RusuRSVPOH19, DBLP:conf/nips/RajeswaranFKL19} and few-shot learning \cite{DBLP:conf/iclr/RaviL17, DBLP:conf/nips/SnellSZ17, DBLP:conf/iclr/SatorrasE18, DBLP:conf/cvpr/SunLCS19, DBLP:conf/nips/ZhangCGBS18, DBLP:conf/nips/OreshkinLL18}. It consists of organising training in a series of learning problems (\textit{a.k.a.}, episodes), each divided into a small labeled training (\textit{`support' set}) and unlabeled validation subset (\textit{`query' set}) to mimic the few-shot circumstances encountered during evaluation. Li \textit{et al}. \cite{DBLP:conf/iccv/LiZYLSH19} proposed to generate episodes to train the model for domain generalization. In each generated episode, the domain-specific model which is mismatched with the current input data will be paired to a domain agnostic module. Qiao \textit{et al}. \cite{DBLP:conf/iccv/Qiao000HW19} tackled the few-shot learning by building up an episodic-wise metric for a series of few-shot tasks. The entire embedding of each task is adapted from a shared task-agnostic feature space into a more discriminative task-specific metric space. The learned classifier can then be effectively leveraged to solve the unseen target classification problem. The advantages of taking the episodic training in this work are two-fold: (1) It allows the source data to be sampled from the class-conditional distribution, so that each target sample can find similar source samples and aggregate the features across the domains. (2) The training episode (Phase 1) and test episode (Phase 2) are symmetric, which facilitates the pseudo labeling as the training and test conditions match well \cite{DBLP:conf/nips/VinyalsBLKW16}.}

\textcolor{black}{
\subsection{Edge-learning Graph Neural Networks}
There is a line of work \cite{DBLP:conf/iclr/Johnson17,DBLP:conf/nips/KimNKY11,DBLP:conf/cvpr/GongC19,DBLP:conf/acml/YangL20,DBLP:conf/cvpr/KimKKY19,DBLP:conf/icml/Brockschmidt20,DBLP:journals/tgrs/ZuoYLZT22} that explores how to update the edge features in graph neural networks for a wide range of tasks. Kim \textit{et al.} \cite{DBLP:conf/nips/KimNKY11} performed higher-order correlation clustering over pairwise superpixel graph for the task of image segmentation. Johnson \cite{DBLP:conf/iclr/Johnson17} extended the idea of gated graph neural networks by learning to construct and modify graphs based on textural input. The edges are updated during the learning by deciding existing edge and potential edge to be added or removed. Yang \textit{et al.} \cite{DBLP:conf/acml/YangL20} incorporated a hierarchical dual-level attention mechanism in the graph embedding model where the node and edge attention layers are alternately learned. Different from prior works in EGNN, our work, for the first time, investigates the EGNN to align the features from the source and target domains automatically and learn a domain-agnostic edge network which can benefit the downstream pseudo-labelling task. }

\section{Preliminaries}
    In this section, we introduce the notations, problem settings, definitions, and theoretical analysis for the tasks of OSDA and SF-OSDA.
    \subsection{Definitions and Problem Settings}
    \begin{definition}\textbf{Closed-set Unsupervised Domain Adaptation (UDA).} Let $\mathcal{P}^s$ and $\mathcal{Q}^t_X$ be the joint probability distribution of the source domain and the marginal distribution of the target domain, respectively. The corresponding label spaces for both domains are equal, \textit{i.e.}, $\mathcal{Y}_s = \mathcal{Y}_t$ = \{1, \ldots, C\}, where $C$ is the number of classes. Given the labeled source data $D_s = \{(x_{s_i}, y_{s_i})\}_{i=1}^{n_s}\sim \mathcal{P}^s$ and the unlabeled target data $D_t = \{x_{t_j}\}_{j=1}^{n_t}\sim\mathcal{Q}^t_{X}$, the aim of UDA is to learn a feature transformation $\mathcal{G}: \mathcal{X} \rightarrow \mathcal{X}'$ and an optimal classifier $h \in \mathcal{H}: \mathcal{X}'\rightarrow \mathcal{Y}$, such that, the learnt model $\mathcal{G} \circ h$ can correctly classify the target samples. $\mathcal{H}$ is the hypothesis space of classifiers, with $n_s$ and $n_t$ indicating the size of source and target dataset, respectively.
    \end{definition}

    \begin{definition}\textbf{Open-set Domain Adaptation (OSDA)~\cite{OSBP}.}
    Different from the UDA setup, OSDA allows the target label space $\mathcal{Y}_t = \{\mathcal{Y}_s, unk\} = \{1, \ldots ,C+1\}$ to include the additional unknown class $C + 1$, which is not present in the source label space $\mathcal{Y}_s$.
    Given independent and identically distributed (\textit{i.i.d.}) samples drawn from the source domain $D_s$ and target domain $D_t$, the goal of OSDA is to train a model $G \circ h$ such that the model can classify the samples from known classes $\{1, \ldots, C\}$ and identify the samples coming from additional unknown class $C + 1$. 
    \end{definition}
    
    \begin{definition}\textbf{Source-free Open-set Domain Adaptation (SF-OSDA).} SF-OSDA aims to adapt the model to the target domain without having access to the source data. Given the model $G_s \circ h_s$ pre-trained on the source set $D_s$, and the unlabeled target set $D_t$, the goal of SF-OSDA is to adapt the model to $G_t \circ h_t$, such that the adapted model is able to correctly classify target samples into the shared classes and the unknown class.
    \end{definition}
    
    \subsection{Risks and Partial Risks}
    \noindent Risks and partial risks are fundamental notions in learning theoretical bounds of OSDA and SF-OSDA. The source risk $\mathfrak{R}_s(h)$ and target risk $\mathfrak{R}_t(h)$ of a classifier $h\in\mathcal{H}$ with respect to the source joint distribution $\mathcal{P}^s$ and the target joint distribution $\mathcal{Q}^t$ are given by,
    \begin{align}
        \mathfrak{R}_s(h) &= \mathbb{E}_{(x, y)\sim\mathcal{P}^s}\mathcal{L}(h(x), y) = \sum_{i=1}^C\pi_{i}^s \mathfrak{R}_{s,i}(h),\nonumber\\
        \mathfrak{R}_t(h) &= \mathbb{E}_{(x, y)\sim\mathcal{Q}^t}\mathcal{L}(h(x), y) = \sum_{i=1}^{C+1}\pi_{i}^t \mathfrak{R}_{t,i}(h),
    \end{align}
    where $\pi_i^s = \mathcal{P}^s(y=i)$ and $\pi_i^t = \mathcal{Q}^t(y=i)$ are class-prior probabilities of the source and target distributions, respectively. The bounded loss function $\mathcal{L}: \mathcal{Y}_t\times \mathcal{Y}_t\rightarrow \mathbb{R}$ satisfies symmetry and triangle inequality. Particularly, the target risk can be split into two partial risks $\mathfrak{R}_{t,*}(h)$ and $\mathfrak{R}_{t,C+1}(h)$, indicating the risks for the known target classes and the unknown class,
    \begin{equation}
        \mathfrak{R}_t(h) = \pi_{C+1}^t \mathfrak{R}_{t,C+1}(h) + (1 - \pi_{C+1}^t)\mathfrak{R}_{t,*}(h),
    \end{equation}
    where the respective partial risks are defined as,
    \begin{equation}
        \begin{split}
           &\mathfrak{R}_{t,*}(h) = \frac{1}{1-\pi_{C+1}^t}\int_{\mathcal{X}'\times Y_s} \mathcal{L}(h(x), y) \mathcal{Q}_{XY|Y_s}(x, y) dx dy,\\
        &\mathfrak{R}_{t,C+1}(h) = \int_{\mathcal{X}'} \mathcal{L}(h(x), C+1) \mathcal{Q}_{X'|C+1}(x) dx.
        \end{split}
    \end{equation}
    \noindent To derive the generalization bound for open-set domain adaptation, we first define a discrepancy measure between the source and target domains:
    \begin{definition}\textbf{{Discrepancy Distance}~\cite{discrepancy}.} For any $h, h' \in \mathcal{H}$, the discrepancy between the distributions of the source and target domains can be formulated as:
    \begin{equation}
        \text{disc}(\mathcal{P}^s, \mathcal{Q}^t) = \sup_{h, h'\in\mathcal{H}}|\mathbb{E}_{\mathcal{P}^s}\mathcal{L}(h, h') - \mathbb{E}_{\mathcal{Q}^t}\mathcal{L}(h, h')|.
    \end{equation}
    Notably, the discrepancy distance is symmetric and satisfy the triangle inequality.
    \end{definition}
    Given the definition of the discrepancy distance, generalization bounds for open-set domain adaptation can be derived as:
    \begin{theorem}\textbf{\textit{OSDA Generalization Bounds}}~\cite{open_theory}.\label{thm:OUDA}
    \label{open-set bound}
    Given the hypothesis space $\mathcal{H}$ with a mild condition that constant vector value function $g:=C + 1\in\mathcal{H}$, $\forall h\in\mathcal{H}$, the expected error on target samples $\mathfrak{R}_t(h)$ is bounded by,
    \begin{equation}\label{eq:OUDA}
        \begin{split}
          \frac{\mathfrak{R}_t(h)}{1-\pi_{C+1}^t}&\leq \overbrace{\mathfrak{R}_s(h)}^{\text{source~risk}} + \overbrace{2disc(\mathcal{Q}^t_{X|Y_s}, \mathcal{P}^s_X)}^{\text{discrepancy~distance}} + \lambda \\
              &+\underbrace{\frac{\mathfrak{R}_{t, C+1}(h)}{1-\pi_{C+1}^t} - \mathfrak{R}_{s, C+1}(h)}_{\text{open~set~risk}~\Delta_{o}},
        \end{split}
    \end{equation}
    \noindent where the shared error $\lambda =\min_{h\in\mathcal{H}}\mathfrak{R}_t^*(h) + \mathfrak{R}_s(h)$. 
    \end{theorem}
    \begin{remark}
    To compute the error upper bound for the closed-set unsupervised domain adaptation, Theorem \ref{thm:OUDA} can be reduced to:
    \begin{equation}
        \begin{split}
            \mathfrak{R}_t(h) &\leq \mathfrak{R}_s(h) +  disc(\mathcal{Q}^t_X, \mathcal{P}^s_X) + \lambda',\\
        \end{split}
    \end{equation}
    with $\pi_{C+1}^t = 0$ and $\lambda' = \min_{h\in\mathcal{H}}\mathfrak{R}_t(h)+\mathfrak{R}_s(h)$.
    \end{remark}
    \noindent According to Eq. \eqref{eq:OUDA}, the target error is bounded by four terms, which opens four directions for improvement:
    \begin{itemize}
        \item Source risk $\mathfrak{R}_s(h)$. A part of the source risk can be avoided based on the assumption that the source domain does not include any unknown samples. This, in turn,  minimizes the upper bound of the error. This direction is rarely investigated in the existing literature of open-set domain adaptation.
    
        \item Discrepancy distance $disc(\mathcal{Q}^t_{X|Y_s}, \mathcal{P}^s_X)$. Minimizing the discrepancy distance between the source and the target domains has been well investigated in recent years in statistics-based~\cite{MMD} or adversarial-based approaches~\cite{DANN}.
        
        \item Shared error $\lambda$ of the joint optimal hypothesis $h^*$. The mismatch in class-wise conditional distributions enlarges the shared error $\lambda$, even when the marginal distributions are aligned.
        
        \item Open set risk $\Delta_{o}$. As shown in Eq. \eqref{eq:OUDA}, the first term of $\Delta_{o}$ can be interpreted as the mis-classification rate for the unknown samples in the target, and the second term is the rate of mis-classifying the source samples as unknown. Therefore, when a large percentage of data is unknown ($\pi_{C+1}^t\rightarrow 1$), open set risk $\Delta_{o}$ contributes the most to the error bound.
        
    \end{itemize}
    
     \begin{figure}[t]
        \centering
        \includegraphics[width=1\linewidth]{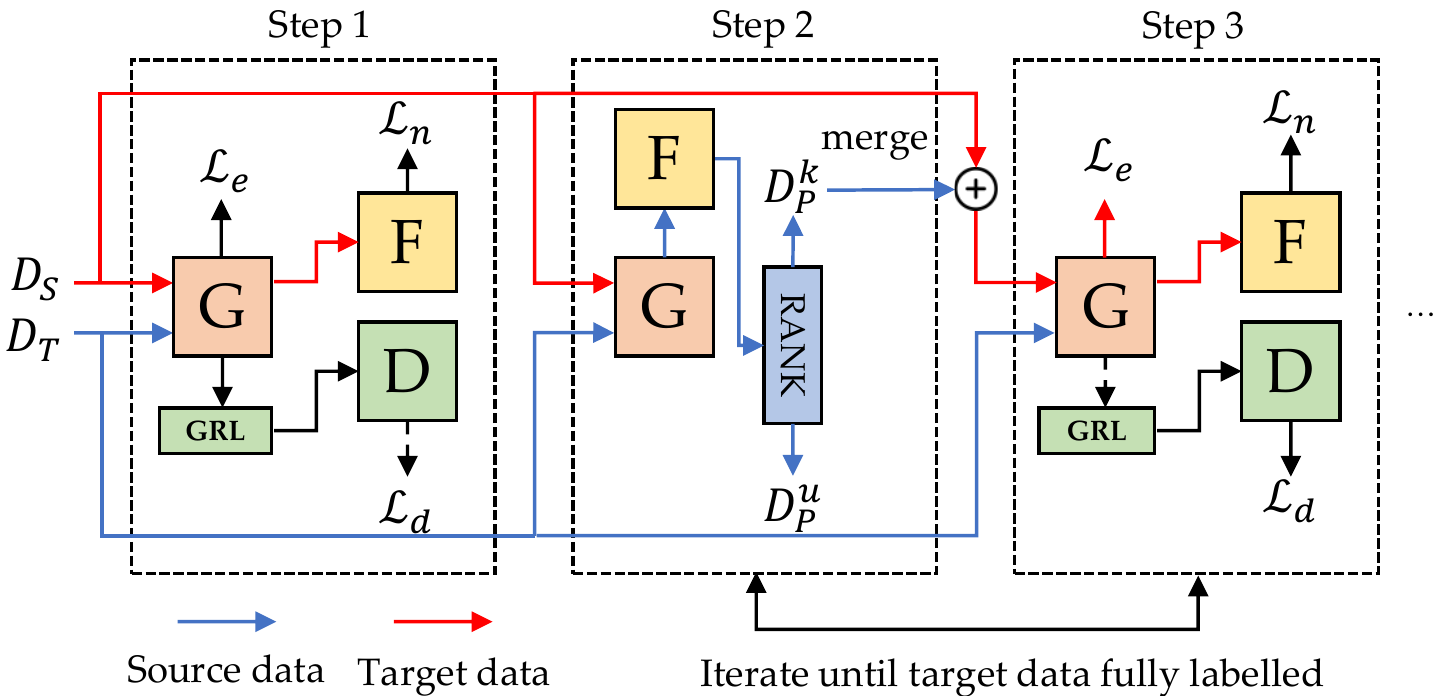}
        \caption{Proposed PGL framework. By alternating between \textit{Steps 2} and \textit{3}, we progressively achieve the optimal classification model $G \circ F$ for the shared classes and pseudo-labeling function $h_b$ for rejecting the unknowns.}
        \label{fig:flowchart}
    \end{figure}
    

    \section{Progressive Graph Learning (PGL)}
        Aiming to minimise the four partial risks mentioned above, we reformulate the open-set unsupervised domain adaptation in a progressive way, and as such, we redefine the task at hand as follows.

    \subsection{Definitions and Risks}
       
    

    \begin{definition}\textbf{Progressive Open-Set Domain Adaptation (POSDA).}\label{def:PGL}
    Given the labeled source data $D_s = \{(x_{s_i}, y_{s_i})\}_{i=1}^{n_s}\sim \mathcal{P}^s$ and unlabeled target data $D_t = \{x_{t_j}\}_{j=1}^{n_t}\sim\mathcal{Q}^t_{X}$, the main goal is to learn an optimal target classifier $\tilde{h}\in \mathcal{H}_1\subset\mathcal{H}$ for the shared classes $Y_s = \{1, \ldots, C\}$ and a pseudo-labeling function $h_b\in\mathcal{H}_2\subset\mathcal{H}$ for the unknown class $C+1$. 
    
    Given the target set will be pseudo-labeled through $M$ steps, the enlarging factor for each step can be defined as $\alpha = \frac{1}{M}$. As long as the hypothesis $\tilde{h}$ and $h_b$ share the same feature extraction part, we can decompose the shared hypothesis $\tilde{h}$ into $\tilde{h}(x) = \argmax_{i\in Y_s}p(i|x)$ and define the pseudo-labeling function $h_b$ at the $m$-th step in line with $\tilde{h}$'s prediction:
    \begin{equation}\label{eq:rank}
        h_b^{(m)} = \begin{cases}
      C+1, & \text{if}~\rank(\max_{i\in Y_s}p(i|x))\leq\alpha_{u}^{(m)}, \\
      \tilde{y}, & \text{if}~\rank(\max_{i\in Y_s}p(i|x))\geq\alpha_{k}^{(m)}.
    \end{cases}
    \end{equation}  $\alpha_u^{(m)} = \beta . \alpha . m . n_t$ and $\alpha_k^{(m)} = n_t - (1-\beta) . \alpha .  m . n_t$ are the index-based thresholds to classify the unknown and known samples, respectively. The hyperparameter $\beta\in(0, 1)$ measures the openness of the given target set as the ratio of unknown samples. $\rank(\cdot)$ is a global ranking function which ranks predicted probabilities in ascending order and returns the sorted index list as an output. The output of pseudo-labeling function is $\tilde{y} = \tilde{h}(x)$ for the possible known samples, and $C+1$ for the unknown ones. 
    \end{definition}
    
    In our case, the upper bound of expected target risk is formulated in the following theorem, 
    \begin{theorem}\textbf{\textit{POSDA Generalization Bound}.}
    \label{thm:POUDA}
    Given the hypothesis space $\mathcal{H}_1$, $\mathcal{H}_2\subset\mathcal{H}$, $\exists \alpha^*\in(0, 1)$, for $\forall \tilde{h}\in\mathcal{H}_1$ and $\forall h_b\in\mathcal{H}_2$, with a condition that the openness $\beta$ of the target set is fixed, the expected error $\mathfrak{R}_t(\tilde{h}, h_b)$ on the target samples is bounded by:

    \begin{equation}
        \begin{split}
          \frac{\mathfrak{R}_t(\tilde{h}, h_b)}{1-\pi_{C+1}^t}&\leq (1-\pi_{\alpha})(\mathfrak{R}_s(\tilde{h}) + disc(\mathcal{Q}^t_{X|Y\leq C}, \mathcal{P}^s_X)) + \lambda\\ &+\underbrace{\frac{\pi_{\alpha}\pi_{C+1}^t}{1-\pi_{C+1}^t}\mathfrak{R}_{t,C+1}(h_b)}_{\text{progressive~open~set~ risk}~\tilde{\Delta}_o} + const,
        \end{split}
    \end{equation}
    \noindent where the shared error $\lambda = \min_{\tilde{h}\in\mathcal{H}_1}\frac{\mathfrak{R}_t^*(\tilde{h})}{1-\pi_{C+1}^t} + (1-\pi_{\alpha})\mathfrak{R}_s(\tilde{h})$. $\pi_{\alpha}$ indicates the prior probability that target samples being pseudo-labeled by $h_b$ (refer to the supplementary material for proof). 
    \end{theorem}
    
    \begin{remark}
     For $\tilde{h}\in\mathcal{H}_1\subset\mathcal{H}$ and $h_b\in\mathcal{H}_2\subset\mathcal{H}$, the following inequality holds, 
     \begin{equation}
        \begin{split}
        \sup_{\tilde{h}\in\mathcal{H}_1} \mathfrak{R}_s(\tilde{h}) &\leq \sup_{h\in\mathcal{H}} \mathfrak{R}_s(h),\\
        \sup_{h_b\in\mathcal{H}_2}\frac{\pi_{C+1}^t}{1-\pi_{C+1}^t} \mathfrak{R}_{t, C+1}(h_b) &\leq \sup_{h\in\mathcal{H}} \frac{\pi_{C+1}^t}{1-\pi_{C+1}^t} \mathfrak{R}_{t, C+1}(h).
        \end{split}
     \end{equation}
     We can observe that our progressive learning framework can achieve a tighter upper bound compared to conventional open-set domain adaptation framework. 
    \end{remark}


    \subsection{Overview}
 
     In this section, we go through the details of the proposed Progressive Graph Learning (PGL) framework, as illustrated in Fig. \ref{fig:flowchart}. Our approach is mainly motivated by the two aspects of minimizing the shared error $\lambda$, and effectively controlling the progressive open-set risk $\tilde{\Delta}_o$ .

\noindent     \textbf{Minimizing the shared error $\lambda$.} \textit{Conditional shift}~\cite{conditionalshift} arises when the class-conditional distributions of the input features substantially differ across the domains, and it is the most significant obstacle for finding an optimal classifier for the source and target data. Specifically, with unaligned distributions of the source distribution $\mathcal{P}_{X|Y}^s$ and target distribution $\mathcal{Q}_{X|Y\leq C}^t$, there is no guarantee to find an optimal classifier for both domains. Therefore, we address the conditional shift in a transductive setting from two perspectives: 
     \begin{itemize}
         \item \textit{Sample-level}: Motivated by~\cite{meta,meta1}, we adopt the episodic training scheme (Section \ref{sec:pgl_step1}), and leverage the source samples from each class to ``support'' predictions on unlabeled data in each episode. While the labeled set is expanding through pseudo-labeling process (Section \ref{sec:progressive}), we progressively update training episodes by replacing the source samples with pseudo-labeled target samples (Section \ref{sec:mixup}).

         \item \textit{Manifold-level}: To regularize the class-specific manifold, we construct $L$-layer Graph Neural Networks (GNNs) on top of the backbone network $G_B(\cdot; \theta_{B})$ (\textit{e.g.}, ResNet). The GNN consists of paired node update networks $G_N(\cdot; \theta_{N})$ and edge update networks $G_E(\cdot; \theta_{E})$. The source nodes and pseudo-labeled target nodes from the same class are densely connected, aggregating information though multiple layers.
     \end{itemize}

\begin{figure}
        \centering
        \includegraphics[width=0.9\linewidth]{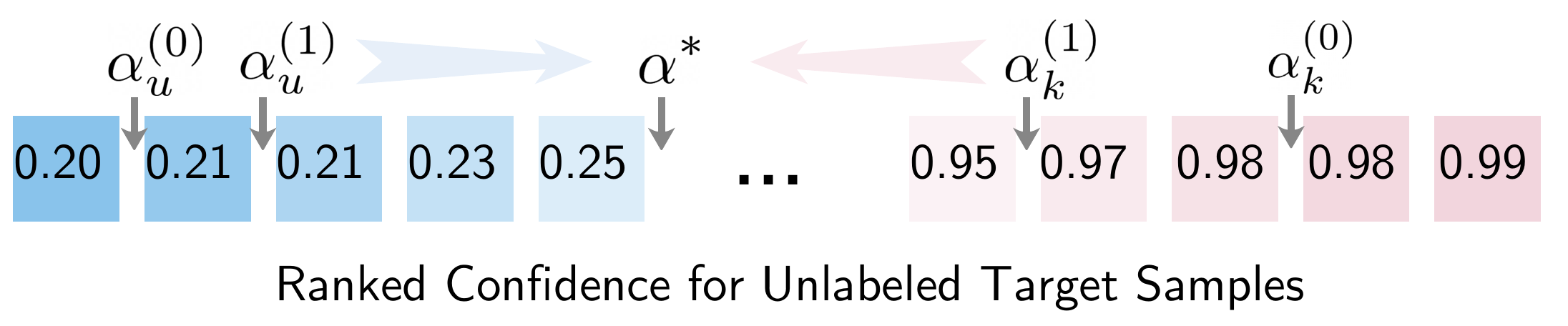}
        \caption{An illustration of the progressive learning to construct the pseudo-labeled target set. $\alpha^*$ indicates the ideal threshold for classifying known and unknown samples.}
        \label{fig:progressive}
    \end{figure}

\noindent     \textbf{Controlling progressive open-set risk $\tilde{\Delta}_o$.} As discussed in Section \ref{sec:progressive}, we iteratively squeeze the index-based thresholds, $\alpha^{(m)}_u$ and $\alpha^{(m)}_k$, to approximate the optimal threshold, $\alpha^*$, as illustrated in Fig. \ref{fig:progressive}. Since the thresholds are mainly determined by the enlarging factor $\alpha$, we can always seek a proper value of $\alpha$ to alleviate the mis-classification error and the subsequent negative transfer. Our experimental results characterize the trade-off between computational complexity and performance improvement.

     \noindent The overall learning procedure can be divided into several steps: (1) \textit{Episodic training with graph neural networks}: the shared classifier $\tilde{h}$ is learned in a transductive setting, along with adversarial objectives for closing domain discrepancy; (2) \textit{Progressive paradigm}: in agreement with $\tilde{h}$'s prediction, all unlabeled target samples are ranked based on confidence, among which we select those with higher scores to form the pseudo-labeled known set and reject ones with lower scores as the unknown set; (3) \textit{Mix-up strategy}: we randomly replace source samples in each episode, with the pseudo-labeled known set obtained from the last step. We will elaborate each of the steps in the next subsections.

    \subsection{Step1: Initial Episodic Training with GNNs}\label{sec:pgl_step1}
    Firstly, we denote the initial episodic formulation of a batch input as $\Tau^{(0)} = \{\Tau_s^{(0)}, \Tau_t^{(0)}\}=\{\tau_{s,i}^{(0)}, \tau_{t,i}^{(0)}\}_{i=1}^B$, with $B$ as the batch size. Each episode in the batch consists of two parts, \textit{i.e.}, the source episode $\tau_{s,i}^{(0)}  = \{(x_i, y)\}_{i=1}^C\sim \mathcal{P}^s_{X|Y}$ randomly sampled from each class $c\in Y_s$ and the target episode $\tau_{t,i}^{(0)} = \{x_j\}_{j=C+1}^{2C}\sim\mathcal{Q}^t_X$ randomly sampled from the target set. All instances in a mini-batch can form an undirected graph $\mathcal{G} = (\mathcal{V}, \mathcal{E})$. Each vertex $v_i\in\mathcal{V}$ is associated with a source or a target feature, and the edge $e_{ij}\in\mathcal{E}$ between nodes $v_i$ and $v_j$ measures the node affinity. \textcolor{black}{The integrated $L$-layer GNNs $G = \{G_B, \{G^{(l)}_E, G^{(l)}_N\}_{l=1}^L\}$} are naturally able to perform a transductive inference taking advantage of labeled source data and unlabeled target data. The propagation rule for edge update and node update is elaborated in the following subsections.

    \noindent\textbf{Edge Update}.
    The generic propagation rule for normalized edge features at the $l$-th layer can be defined as,
    \begin{align}\label{eq:adjacency}
            A_{ij}^{(l)} &= \sigma \big(G_E^{(l)}(\norm{v_i^{(l-1)} - v_j^{(l-1)}}; \theta_E^{(l)})\big),\nonumber\\ 
            \mathcal{E}^{(l)} &= D^{-\frac{1}{2}}(A^{(l)} + I) D^{-\frac{1}{2}},
    \end{align}
    with $\sigma$ being the sigmoid function, $D$ the degree matrix of $A^{(l)} + I$, $I$ the identity matrix, and $G_E^{(l)}(\cdot; \theta_E^{(l)})$ the non-linear edge network parameterized by $\theta_E$. 
    
    \begin{figure}
        \centering
        \includegraphics[width=0.85\linewidth]{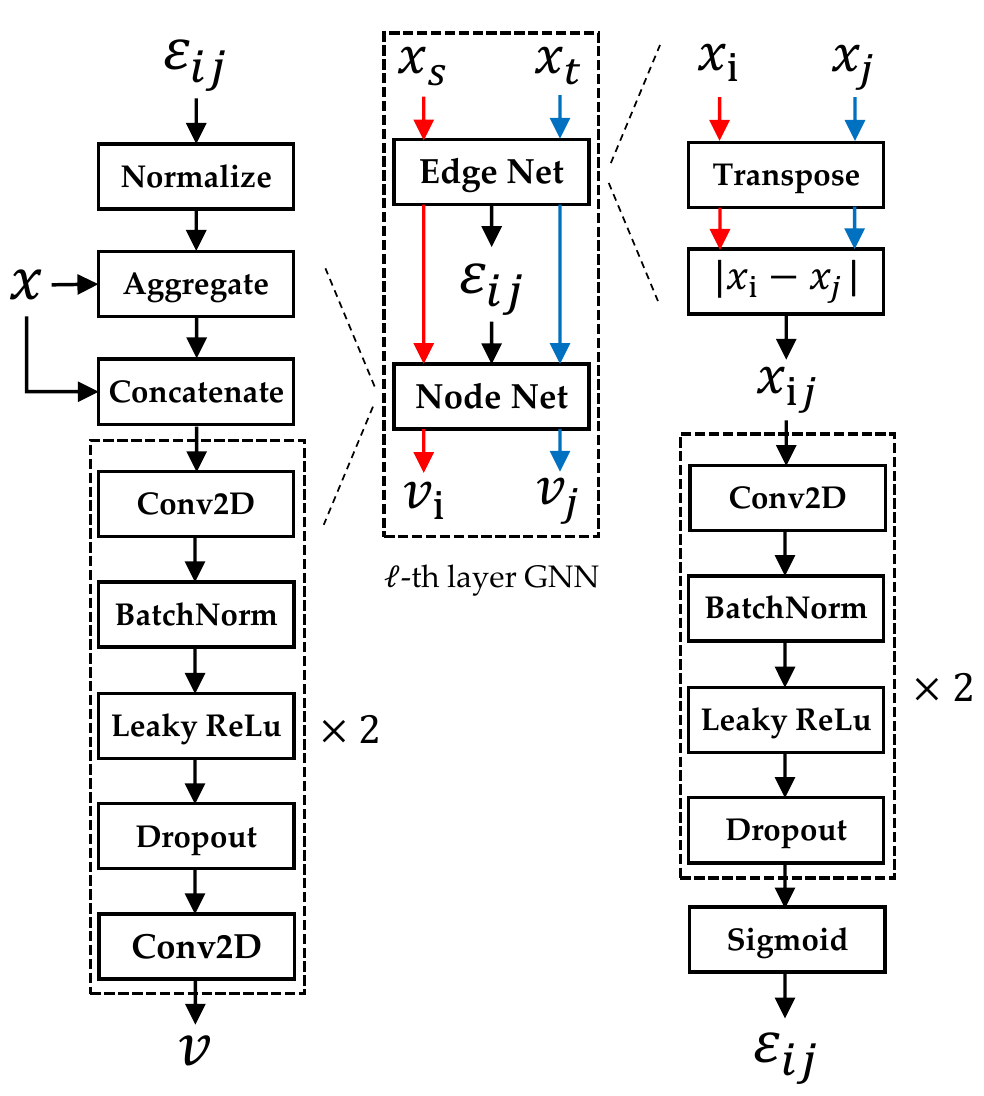}
        \caption{The network architecture of the node network $G_N$ and edge network $G_E$.}
        \label{fig:gnn}
    \end{figure}
    \noindent\textbf{Node Update}.
    Similarly, the  propagation rule for node features at the $l$-layer is defined as,
    \begin{align}\label{eq:node}
        &\hat{v}_i^{(l-1)}= \sum_{j\in\mathcal{N}(i)}(v_i^{(l-1)} \mathcal{E}^{(l-1)}_{ij}),\nonumber\\
        &v_i^{(l)} = G_N^{(l)}([v_i^{(l-1)}; \hat{v}_i^{(l-1)}];\theta_N^{(l)}),
    \end{align}
    with $\mathcal{N}(i)$ being the neighbor set of the node $v_i$, $[\cdot;\cdot]$ the concatenation operation and $G_N^{(l)}(\cdot;\theta_N^{(l)})$ the node network consisting of two convolutional layers, LeakyReLU activations and dropout layers. The node embedding is initialized with the extracted representations from the backbone embedding model, \textit{i.e.}, $v_i^{(0)}=G_B(x_i)$.

   \noindent\textbf{Adaptive Learning.} We exploit adversarial loss to align the distributions of the source and target features extracted from the backbone network $G_B(\cdot;\theta_B)$. Specifically, a domain classifier $D(\cdot; \theta_D)$ is trained to discriminate between the features coming from the source or target domains, along with a generator $G_B$ to fool the discriminator $D$. The two-player minimax game shown in~Eq.\eqref{Eq:minmax} is expected to reach an equilibrium resulting in the domain invariant features:
    \begin{equation}
        \begin{split}
        \mathcal{L}_d = \mathbb{E}_{x\sim\Tau_s}\log [D(G_B(x))]+\mathbb{E}_{x\sim\Tau_t}\log [1 - D(G_B(x))].    
        \end{split}
        \label{Eq:minmax}
    \end{equation}
    \textbf{Node Classification.} By decomposing the shared hypothesis $\tilde{h}$ into a feature learning module $G(\cdot,\theta_G)$ and a shared classifier $F(\cdot, \theta_F)$, we train both networks to classify the source node embedding. To alleviate the inherent class imbalance issue, we adopt the focal loss to down-weigh the loss assigned to correctly-classified examples:
    \begin{equation}
        \mathcal{L}_n = -\mathbb{E}_{(x,y)\sim\Tau_s}\sum_{l=1}^{L}(1-F(G(x)^{(l)}))^\rho\log[F(G(x)^{(l)})],
    \end{equation}
    with the hyperparameter $\rho = 2$ and $G(x)^{(l)}$ being the node embedding from the $l$-th node update layer. The total loss combines all losses from $L$ layers to improve the gradient flow in the lower layers.

\noindent\textbf{Edge Classification.} Based on the given labels of the source data, we construct the ground-truth of edge map $\widehat{Y}$, where $\widehat{Y}_{ij} = 1$ if $x_i$ and $x_j$ belong to the same class, and $\widehat{Y}_{ij} = 0$, otherwise. The networks are trained by minimizing the following binary cross-entropy loss:
    \begin{equation}
        \mathcal{L}_e = -\mathbb{E}_{(x,y)\sim\Tau_s}\sum_{l=1}^{L} \widehat{Y} \log \mathcal{E}^{(l)} + (1-\widehat{Y})\log[1-\mathcal{E}^{(l)}].
    \end{equation}
    
    \noindent\textbf{Final Objective Function.} Formally, our ultimate goal is to learn the optimal parameters for the proposed model,
    \begin{equation}\label{eq:optimization}
    \begin{split}
        (\theta^*_N, \theta^*_E, \theta^*_F, \theta^*_D) = \argmin \mathcal{L}_n + \mu\mathcal{L}_e + \gamma\mathcal{L}_d,\\
        (\theta^*_B) = \argmin \mathcal{L}_n + \mu\mathcal{L}_e - \gamma\mathcal{L}_d,
    \end{split}
    \end{equation}
    with $\mu$ and $\gamma$ the coefficients of the edge loss and adversarial loss, respectively.

    \subsection{Step2: Pseudo-Labeling in Progressive Paradigm}\label{sec:progressive}
    With the optimal model parameters obtained at the $m$-th step, we freeze the model and feed forward all the target samples, as shown in the \textit{Step 2} of Fig. \ref{fig:flowchart}. Then, we rank the maximum likelihood $\max_{i\in Y_s}p(i|x)$ produced from the shared classifier $F(G(x)^{(L)})$ in ascending order. Giving priority to the ``easier'' samples with relatively high/low confidence scores, we select $\alpha . m . n_t$ samples to enlarge the pseudo-labeled known set $\mathcal{D}_P^k$ and unknown set $\mathcal{D}_P^u$, \textcolor{black}{where $.$ indicates the scalar multiplication}:
    \begin{equation}
        \begin{split}
            \mathcal{D}_P^k\leftarrow\mathcal{D}_k^{(0)}\cup\mathcal{D}_k^{(1)}\ldots\cup\mathcal{D}_k^{(m)},\\
            \mathcal{D}_P^u\leftarrow\mathcal{D}_u^{(0)}\cup\mathcal{D}_u^{(1)}\ldots\cup\mathcal{D}_u^{(m)},\\
            \mathcal{D}_k^{(m)} = \{(x_i, \tilde{y}_i)\}_{i=1}^{(1-\beta) . \alpha . m . n_t},\\
            \mathcal{D}_u^{(m)} = \{(x_j, C+1)\}_{j=1}^{\beta . \alpha . m. n_t}.
        \end{split}
    \end{equation}
    $\mathcal{D}_k^{(m)}$ and $\mathcal{D}_u^{(m)}$ are newly annotated known set and unknown set, respectively, and the pseudo-label is given by $\tilde{y}_i = \argmax_{y\in Y_s}p(y|x)$. To find a proper value of enlarging factor $\alpha$, we have two options: by aggressively setting a large value to $\alpha$, the progressive paradigm can be accomplished in fewer steps resulting in potentially noisy and unreliable pseudo-labeled candidates; on the contrary, choosing a small value of $\alpha$ can result in a steady increase of the model performance and the computational cost.
    
    \subsection{Step3: Episodic Update with Mix-up Strategy}\label{sec:mixup}
    We mix the source data with the samples from the updated pseudo-labeled known-set $\mathcal{D}^k_P$ at the $m$-th step, and construct new episodes $\Tau^{(m+1)}$ at the $(m+1)$-th step, as depicted in the \textit{Step 3} of Fig. \ref{fig:flowchart}. In particular, We randomly replace the source samples with pseudo-labeled known data with a probability $\mathcal{P}^{(m)}_r=m\alpha$. Each episode in the new batch consists of three parts,
    \begin{equation}
    \begin{split}
        \tau_{s,i}^{(m+1)} &= \{(x_i, y_i)\}_{i=1}^{C\times(1-\mathcal{P}^{(m)}_r)}\sim \mathcal{P}^s(x|y),\\
        \widetilde{\tau}_{t,i}^{(m+1)} &= \{(x_k, \tilde{y}_k)\}_{k=1}^{C\times\mathcal{P}^{(m)}_r}\sim \mathcal{Q}^t(x|\tilde{y}), \\
        \tau_{t,i}^{(m+1)} &= \{x_j\}_{j=C+1}^{2C}\sim\mathcal{Q}^t_X,
    \end{split}
    \end{equation}
    with $\mathcal{Q}^t(x|\tilde{y})$ being the conditional distribution of the pseudo-labeled known set at the $m$-th step. Then, we update the model parameters according to Eq. \eqref{eq:optimization} and repeat pseudo-labeling with the newly constructed episodes until convergence.

\subsection{Extension to Video Domain Adaptation}
We provide an extension of the proposed PGL approach for tasks of open-set video domain adaptation (OSVDA), where target video data is curated under a different condition and contains additional classes of actions or events that do not exist in the source domain. For instance, in the Gameplay-Kinetics \cite{DBLP:conf/iccv/ChenKAYCZ19} dataset, our OSVDA differs from vanilla OSDA in a sense that the domain shift is present in video clips rather than still images. Specifically, we sample a fixed-number of frames, $K$, with an equal spacing from each video for training. We then encode each frame with the Resnet-101 pretrained on ImageNet into a 2048-D vector. Without loss of generality, the extracted frame features for source and target domains are then aggregated through the average pooling layer to obtain the video-level representations. Likewise, the graph-based model $G \circ \tilde{h}$ is jointly trained to align the source and target features, as illustrated in Section \ref{sec:pgl_step1} to Section \ref{sec:mixup}. 

We also provide an extension of the PGL method in a semi-supervised setting for video data (S-OSVDA), where part of the target labels are observable. To leverage the supervision, the node objective $\mathcal{L}_{n}$ and edge loss $\mathcal{L}_{e}$ are adapted to take both source samples and the labeled target samples, while the unlabeled target videos are pseudo-labeled iteratively.

\begin{figure}[t]
        \centering
        \includegraphics[width=1\linewidth]{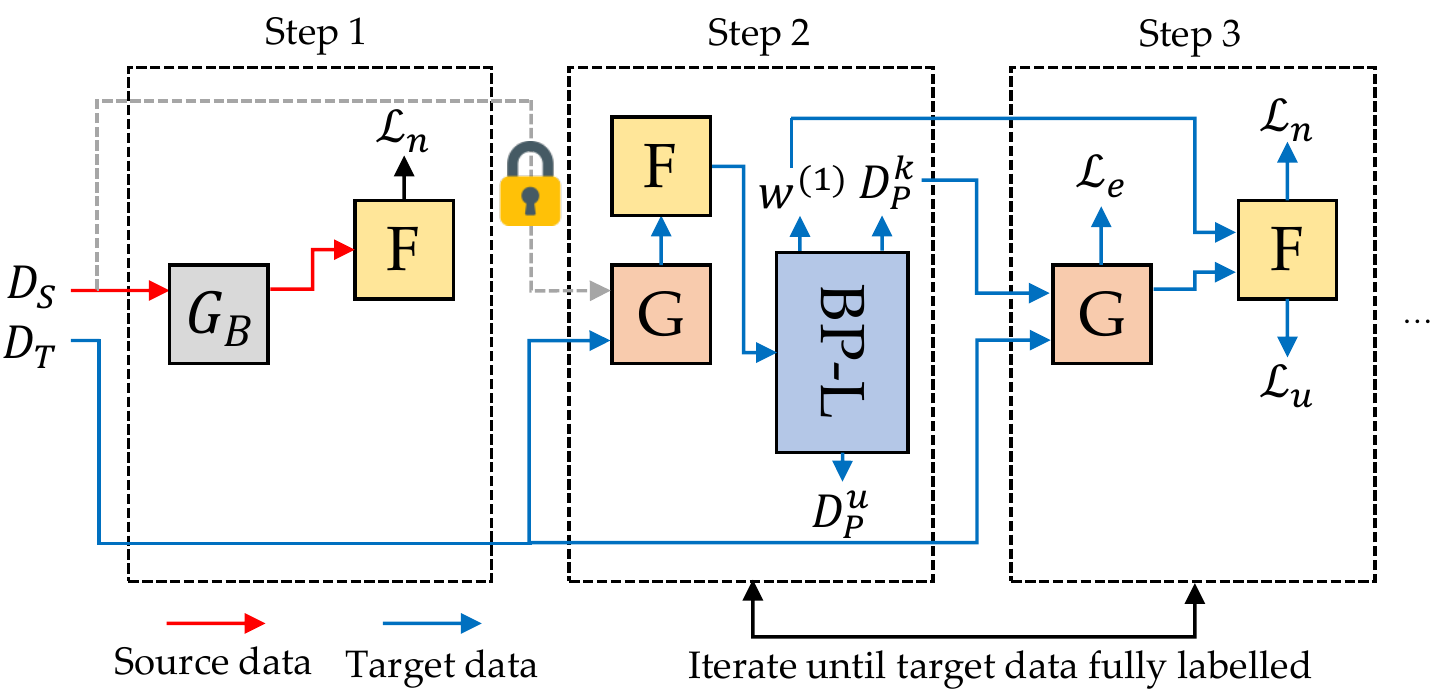}
        \caption{Proposed SF-PGL framework. The black line represents the data flow of both the source and target data. }
        \label{fig:SF_flowchart}
    \end{figure}
\section{Source-free Progressive Graph Learning (SF-PGL)}
In this section, we propose simple yet effective modifications to make our progressive graph learning model work in a source-free setup (SF-PGL). The overall workflow is presented in Fig. \ref{fig:SF_flowchart}, which consists of (1) Pre-training the backbone Model, (2) Balanced pseudo-labeling, and (3) Uncertainty-aware updating.

\subsection{Step1: Pre-training Backbone Model}
 First, we train the backbone network $G_B(:, \theta_B)$ and the source hypothesis module $F(:, \theta_{F})$ on the source domain, using the cross-entropy objective, 
 \begin{equation}
        \mathcal{L}_c = -\mathbb{E}_{(x,y)\sim\Tau_s}\log[F_s(G_B(x))].
\end{equation}
Due to the absence of target samples, the training of the graph model is not involved in Step 1, as otherwise, the model overfits the source data.
\subsection{Step2: Balanced Pseudo-Labeling (BP-L)}
After the backbone model $G_B$ and classifier $F$ is warmed up, we pass all target samples to get the predictions $p(y|x) = F(G_B(x))$ for pseudo labeling. With no access to the labeled source data, we cannot simply apply the global ranking strategies in Section \ref{sec:progressive} to get the pseudo-labeled known set and the unknown set. This is because the global ranking may be biased to some `easy' classes where samples tend to have high confidence, which can cause downsampling of `difficult' classes. The imbalance existed in the pseudo-labeled set is prone to trigger the overfitting for certain categories and notorious overconfidence issue \cite{DBLP:conf/icml/GuoPSW17}, which makes the model poorly calibrated and less generalizable. To be specific, overconfidence refers to the problem that the produced confidence scores are typically higher than the predictive accuracy, which may increase the risks for decision-critical applications. 

To solve this, in SF-PGL framework, we adopt a \textit{balanced pseudo-labeling} (BP-L) mechanism in the inference stage. Firstly, we separate all target data into $C$ groups according to their potential labels $\tilde{y} = \argmax_{y\in Y_s}p(y|x)$ and sort the samples based on their confidence scores. For each shared class $c\in C$, we create an empty label bank $\mathcal{B}^{(m)}_c$ of the size $\alpha . m . n_t / C$ and then insert the highest confidence scores $\{\hat{p}(y|x) | x\sim \mathcal{P}(x|\tilde{y}=c)\}$ from the $c$-th group into the label bank until fully filled. Afterwards, we merge $C$ label banks to form the pseudo-labeled known set $\mathcal{D}_P^k$ at the $m$-th step:
 \begin{equation}
            \mathcal{D}_k^{(m)} = \{(x_i, \tilde{y}_i) | p(y_i|x_i) \in \mathcal{B}^{(m)}_c\}_{c=1}^{C}.
    \end{equation}
The pseudo-labeled unknown set $\mathcal{D}_P^u$ is collected in the same way described in Section \ref{sec:progressive}, by obtaining the target samples with the lowest maximum confidence scores, such as
\begin{equation}
            \mathcal{D}_u^{(m)} = \{(x_j, C+1)\}_{j=1}^{\beta . \alpha . m. n_t}.
\end{equation}
In order to measure the class uncertainty, we further record the confidence threshold $\gamma_c$ in each label bank and concatenate the thresholds as $\gamma^{(m)}$,
\begin{equation}
    \gamma^{(m)} = [\gamma^{(m)}_1, \gamma^{(m)}_2, \ldots, \gamma^{(m)}_C], \gamma_c^{(m)} = \min_{\hat{p}\in \mathcal{B}^{(m)}_c} \hat{p}(y| x),
\end{equation}
where $[\cdot,\cdot]$ denotes the concatenation. Then we normalize the confidence thresholds to calculate the class importance $w^{(m)}\in\mathbb{R}^C$ at the $m$-th step,
\begin{equation}
    w^{(m)} = \texttt{Softmax}(1 - \gamma^{(m)}),
\end{equation}
where $\texttt{Softmax}(x) = \frac{e^{x_i}}{\sum_{j=1}^C e^{x_j}}$ is the standard softmax function. The value of $w^{(m)}$ varies from 0 to 1. A smaller value indicates the class is relatively easy to learn, whilst a larger value means the class is difficult and should pay more attention to learn the concept. 

\subsection{Step3: Uncertainty-aware Updating}
With the pseudo-labeled known set constructed, we leverage the same episodic training paradigm used in Section \ref{sec:pgl_step1} to train the backbone $G_B$, the node and edge network $G_N$ and $G_E$ and the classifier $F$. At the $(m+1)$-th step, each episode in the new batch consists of labeled data sampled from $\mathcal{D}_k^{(m)}$ and unlabeled target data sampled i.i.d.,
\begin{equation}
    \begin{split}
        \widetilde{\tau}_{t,i}^{(m+1)} &= \{(x_k, \tilde{y}_k)\}_{k=1}^{C}\sim \mathcal{Q}^t(x|\tilde{y}), \\
        \tau_{t,i}^{(m+1)} &= \{x_j\}_{j=C+1}^{2C}\sim\mathcal{Q}^t_X.
    \end{split}
\end{equation}

The networks are trained by minimizing the edge loss $\mathcal{L}_e$, supervised node classification loss $\mathcal{L}_n$ and a soft entropy loss $\mathcal{L}_u$ for unlabeled target data as defined below,
\begin{equation}
        \begin{split}
        &\textcolor{black}{(\theta^*_N, \theta^*_E, \theta^*_F, \theta^*_B) = \argmin \mathcal{L}_n +  \mu\mathcal{L}_e,}\\
        &\mathcal{L}_n = -\mathbb{E}_{(x,y)\sim\widetilde{\Tau}_t}\sum_{l=1}^{L} w^{(m)}\log[F(G(x)^{(l)})],\\
        &\mathcal{L}_e = -\mathbb{E}_{(x,y)\sim\widetilde{\Tau}_t}\sum_{l=1}^{L} \widehat{Y} \log \mathcal{E}^{(l)} + (1-\widehat{Y})\log[1-\mathcal{E}^{(l)}],
        \end{split} \vspace{-5ex}
        \nonumber
    \end{equation}
 Different from objectives in PGL, the node loss is weighted by the $w^{(m)}$ obtained from the last step, hereby the learning for difficult categories can be enhanced. This strategy also alleviates the issues of overfitting and miscalibration evidenced by experimental results shown in Section \ref{sec:exp}. Finally, we update the model parameters and repeat balanced pseudo-labeling with the newly constructed episodes until convergence.
\section{Experiments}
In this section, we quantitatively compare our proposed model against various domain adaptation baselines on three image classification and four action recognition datasets. 


\subsection{Datasets}
To testify the versatility, we evaluate the proposed PGL and SF-PGL methods over three image recognition and four action recognition benchmarks as introduced below.

\textbf{Office-Home}~\cite{officehome} is a challenging domain adaptation benchmark, which comprises 15,500 images from 65 categories of everyday objects. The dataset consists of 4 domains: Art (\textbf{Ar}), Clipart (\textbf{Cp}), Product (\textbf{Pr}), and Real-World (\textbf{Rw}). Following the same splits used in ~\cite{STA}, we select the first 25 classes in alphabetical order as the known classes, and group the rest of the classes as the unknown. 

\textbf{VisDA-17}~\cite{visda2017} is a cross-domain dataset with 12 categories in two distinct domains. The \textbf{Synthetic} domain consists of 152,397 synthetic images generated by 3D rendering and the \textbf{Real} domain contains 55,388 real-world images from MSCOCO~\cite{MSCOCO} dataset. Following the same protocol used in ~\cite{OSBP, STA}, we construct the known set with 6 categories and group the remaining 6 categories as the unknown set. 

\textbf{Syn2Real-O}~\cite{visda18} is the most challenging synthetic-to-real testbed, which is constructed from the \textit{VisDA-17}. The Syn2Real-O dataset significantly increases the openness to 0.9 by introducing additional unknown samples in the target domain. According to the official setting, the \textbf{Synthetic} source domain contains training data from the \textit{VisDA-17} as the known set, and the target domain \textbf{Real} includes the test data from the \textit{VisDA-17} (known set) plus 50k images from irrelevant categories of MSCOCO dataset (unknown set). 

\begin{table*}[t]
\centering 
\caption{The general statistics of the four action recognition datasets for tasks of OSVDA and S-OSVDA.}
\resizebox{1\textwidth}{!}{
\begin{tabular}{l | c c c c } 
\toprule 
\textbf{Property} &\textbf{UCF-HMDB$_{small}$} &\textbf{UCF-HMDB$_{full}$} &\textbf{UCF-Olympic} &\textbf{Kinetics-Gameplay} \\ 
\midrule 
Video Length & $\sim$21 Seconds & $\sim$33 Seconds &$\sim$39 Seconds &$\sim$ 10 Seconds\\ 
Classes &5 &12 &6 &30\\ 
Training Videos &UCF: 482 / HMDB: 350 &UCF:1,438 / HMDB: 840 &UCF: 601 / Olympic: 250 & Kinetics: 43,378 / Gameplay: 2,625 \\ 
Validation Videos &UCF: 189 / HMDB: 571 &UCF: 360 / HMDB: 350 &UCF: 240 / Olympic: 54 & Kinetics: 3,246 / Gameplay: 749 \\ 
Known/Unknown Classes &4/1 &6/6 &5/1 &15/15\\
\bottomrule 
\end{tabular}
}
\label{tab:video_datasets}
\end{table*}

\begin{table}[t]
\centering
\caption{The summary of all collected categories in the \textit{UCF-101} and \textit{HMDB} datasets. Classes highlighted in blue represent the unknown class.}
\resizebox{0.5\textwidth}{!}{
\begin{tabular}{c|c|c}
\toprule
UCF-HMDB $_{\text {full }}$ & UCF & HMDB \\
\hline climb & RockClimbingIndoor, RopeClimbing & climb \\
\hline fencing & Fencing & fencing \\
\hline golf & GolfSwing & golf \\
\hline kick\_ball & SoccerPenalty & kick\_ball \\
\hline pullup & PullUps & pullup \\
\hline punch & Punch, & punch \\
& BoxingPunchingBag, BoxingSpeedBag & \\
\hline \cellcolor{blue!25}pushup & PushUps & pushup \\
\hline \cellcolor{blue!25}ride\_bike & Biking & ride\_bike \\
\hline \cellcolor{blue!25}ride\_horse & HorseRiding & ride\_horse \\
\hline \cellcolor{blue!25}shoot\_ball & Basketball & shoot\_ball \\
\hline \cellcolor{blue!25}shoot\_bow & Archery & shoot\_bow \\
\hline \cellcolor{blue!25}walk & WalkingWithDog & walk \\
\bottomrule
\end{tabular}
}
\label{tab:class_ucfhmdb}
\end{table}

We adapted four video DA benchmark datasets for the open-set domain adaptation setting by resplitting the label spaces, of which the statistics are summarized in Table \ref{tab:video_datasets}. For each source and target video, we sample a fixed number $K$ with equal spacing and encode each frame with the ResNet-101 pretrained on ImageNet into a 2048-D vector. In our experiments, $K$ is empirically set to 5 for all OSVDA approaches.

The \textbf{UCF-HMDB$_{small}$} and \textbf{UCF-HMDB$_{full}$} are the overlapped subsets of two large-scale action recognition datasets, \textit{i.e.}, the UCF101~\cite{ucf} and HMDB51~\cite{hmdb}, covering 5 and 12 highly relevant categories respectively. Each category may correspond to multiple categories in the original UCF101 or HMDB51 dataset, as shown in Table \ref{tab:class_ucfhmdb}. \textbf{UCF-HMDB$_{small}$} only contains \textit{golf, pullup, ride\_bike, ride\_horse, shoot\_ball}, where the \textit{shoot\_ball} is considered as the unknown class. For \textbf{UCF-HMDB$_{full}$}, the classes highlighted in blue (Table \ref{tab:class_ucfhmdb}) are grouped as the unknown class. 

The \textbf{UCF-Olympic} selects the shared 6 classes from the UCF101 and Olympic Sports Datasets~\cite{olympic}, including \textit{Basketball, Clearn and Jerk, Diving, Pole Vault, Discus Throw and Tennis}, where the \textit{Tennis} acts as the unknown class.

\textbf{Kinetics-Gameplay}: The fourth and most challenging dataset is the cross-domain Kinetics-Gameplay dataset, which has a large domain gap between its synthetic videos and real-world videos. To create this dataset, 30 shared categories were selected from both the Gameplay~\cite{DBLP:conf/iccv/ChenKAYCZ19} dataset and one of the largest public video datasets, Kinetics-600: \textit{break, carry, clean floor, climb, crawl, crouch, cry, dance, drink, drive, fall down, fight, hug, jump, kick, light up, news anchor, open door, paint brush, paraglide, pour, push, read, run, shoot gun, stare, talk, throw, walk, and wash dishes}. Each category in Kinetics-Gameplay may also correspond to multiple categories in both datasets, which poses another challenge of class imbalance. We manually select the last 15 classes as the unknown class.

\subsection{Baselines}
We compare the performance of the proposed PGL and SF-PGL models with 1) a basic \textbf{ResNet-50}~\cite{resnet} deep classification model; 2) closed-set domain adaptation methods:  Maximum Mean Discrepancy (\textbf{MMD}) \cite{MMD}, Domain-Adversarial Neural Networks (\textbf{DANN}) \cite{DANN}, Residual Transfer Networks (\textbf{RTN}) \cite{RTN}, Joint Adaptation Networks (\textbf{JAN}) \cite{JAN}, Maximum Classifier Discrepancy (\textbf{MCD}) \cite{MCD}, 3) partial domain adaptation methods: Adaptive Batch Normalization (AdaBN) \cite{AdaBN}, Importance Weighted Adversarial Nets (\textbf{IWAN}) \cite{IWAN}, Example Transfer Network (\textbf{ETN}) \cite{ETN} 4) open-set domain adaptation methods: Assign-and-Transform-Iteratively (\textbf{ATI-$\lambda$}) \cite{ATI}, Open Set domain adaptation by Back-Propagation (\textbf{OSBP}) \cite{OSBP}, \textbf{STA}\cite{STA}, \textbf{DAOD}~\cite{open_theory}, \textbf{ROS}\cite{DBLP:conf/eccv/BucciLT20}, Self-Ensembling
with Category-agnostic Clusters (\textbf{SE-CC}) \cite{DBLP:conf/cvpr/PanYLNM20}, 5) universal domain adaptation method (UAN) \cite{DBLP:conf/cvpr/YouLCWJ19} and 6) source-free open-set domain adaptation approach \textbf{Inherit} \cite{DBLP:conf/cvpr/KunduVRVB20}. To be able to apply the non-open-set baseline methods in the open-set setting, we follow the previous baselines~\cite{STA,Attract-UTS} and reject unknown outliers from the target data using \textbf{OSVM}~\cite{OSVM} and \textbf{OSNN}~\cite{OSNN}.

\begin{table*}[!t]
	\begin{center}
		\caption{Recognition accuracies (\%) on 12 pairs of source/target domains from \textit{Office-Home} benchmark using ResNet-50 as the backbone. \textbf{Ar}: Art, \textbf{Cp}: Clipart, \textbf{Pr}: Product, \textbf{Rw}: Real-World. $^*$ indicates our re-implementation with the officially released code.}\label{tab:home} \vspace{1ex}
		\resizebox{1\textwidth}{!}{
		\begin{tabular}{lcccccccccccccccccccccccccc}
			\toprule
			\multirow{2}{*}{Method}&
			\multicolumn{2}{c}{\textbf{Ar$\to$Cl}}&
			\multicolumn{2}{c}{\textbf{Ar$\to$Pr}}&
			\multicolumn{2}{c}{\textbf{Ar$\to$Rw}}&
			\multicolumn{2}{c}{\textbf{Cl$\to$Rw}}&
			\multicolumn{2}{c}{\textbf{Cl$\to$Pr}}&
			\multicolumn{2}{c}{\textbf{Cl$\to$Ar}}&
			\multicolumn{2}{c}{\textbf{Pr$\to$Ar}}&
			\multicolumn{2}{c}{\textbf{Pr$\to$Cl}}&
			\multicolumn{2}{c}{\textbf{Pr$\to$Rw}}&
			\multicolumn{2}{c}{\textbf{Rw$\to$Ar}}&
			\multicolumn{2}{c}{\textbf{Rw$\to$Cl}}&
			\multicolumn{2}{c}{\textbf{Rw$\to$Pr}}&
			\multicolumn{2}{c}{\textbf{Avg.}}\\
			\cmidrule(l){2-3}
			\cmidrule(l){4-5}
			\cmidrule(l){6-7}
			\cmidrule(l){8-9}
			\cmidrule(l){10-11}
			\cmidrule(l){12-13}
			\cmidrule(l){14-15}
			\cmidrule(l){16-17}
			\cmidrule(l){18-19}
			\cmidrule(l){20-21}
			\cmidrule(l){22-23}
			\cmidrule(l){24-25}
			\cmidrule(l){26-27}
			& OS & OS$^*$& OS & OS$^*$& OS & OS$^*$& OS & OS$^*$& OS & OS$^*$& OS & OS$^*$& OS & OS$^*$& OS & OS$^*$& OS & OS$^*$& OS & OS$^*$& OS & OS$^*$& OS & OS$^*$& OS & OS$^*$\\
			\midrule
			OSNN~\cite{OSNN} &33.7&32.1 &40.6 &39.4 &57.0 &56.6 &47.7 &46.9 &40.3 &39.1 &34.0 &32.3& 39.7 &38.5 &36.3&35.0 &59.7&59.6 & 52.1 & 51.4 &39.2 & 38.0 &59.2&59.2 & 45.0 & 44.0 \\
			OSVM~\cite{OSVM} &37.5&38.7 &42.2 &42.6 &49.2 &51.4 &53.8 &55.5 &48.5 &50.0 & 39.2 &40.3 &53.4 &55.1 &43.5 &44.8 &70.6 &72.9 &65.6 &67.4 &49.5 & 50.8 &72.7 &75.1 &52.1 &53.7\\
			DANN$^\dagger$\cite{DANN} &52.3 &52.1 &71.3 &72.4 &82.3 &83.8 &73.2 &74.5 &62.8 &64.1 &61.4 &62.3 &63.5 &64.5 &46.0 &46.3 &77.2 &78.3 &70.5 &71.3 &55.5 &56.2 &79.1 &80.7 &66.2 &67.2\\
			ATI-$\lambda^\dagger$\cite{DBLP:journals/pami/BustoIG20} &53.1 &54.2 &68.6 &70.4 &77.3 &78.1 &74.3 &75.3 &66.7 &68.3 &57.8 &59.1 &61.2 &62.6 &53.9 &54.1 &79.9 &81.1 &70.0 &70.8 &55.2 &55.4 &78.3 &79.4 &66.4 &67.4\\
			\midrule
			OSBP\cite{OSBP} &56.1 &57.2& 75.8 & 77.8 &83.0 &85.4 &75.5 &77.2 &69.2 &71.3 &64.6 &65.9 &64.6 &65.3 &48.3 &48.7 &79.5 &81.6 &72.1 &73.5 &54.3 &55.3 &80.2 &81.9 &68.6  &70.1 \\
			STA\cite{STA} &58.1 &- & 71.6 &- &85.0 &- &75.8 &- &69.3 &- &63.4 &- &65.2 &- &53.1 &- &80.8 &- &74.9 &- &54.4 &- &81.9 &- &69.5&-\\
			STA$^*$ &46.6&45.9&67.0&67.2&76.2&76.6&64.9&65.2&57.7&57.6&50.2&49.3&49.5&48.4&42.9&40.8&76.6&77.3&68.7&68.6&46.0&45.4&73.9&74.5&60.0&59.8\\
			ROS\cite{DBLP:conf/eccv/BucciLT20} &51.5 &50.6 &68.5 &68.4 &75.9 &75.8 &65.6 &65.3 &60.3 &59.8 &54.1	&53.6 &57.6	&57.3 &46.5	&46.5 &71.1	&70.8 &67.1	&67.0 &52.3	&51.5 &72.3	&72.0 &62.0	&61.6\\
			DAOD\cite{open_theory} & 56.1 & 55.5 &69.1 & 69.2 &78.7&79.3 & 77.3& 78.2& 69.6& 70.2 &62.6 &62.9 &66.8 &67.7 &\textbf{59.7} & \textbf{60.3}& \textbf{83.3} &\textbf{85.0} & 72.3 &73.2 & 59.9 & 60.4 & 81.8 & 82.8 &69.8 &70.4\\
			\midrule
			\midrule
			\textbf{PGL}&\textbf{61.6} &\textbf{63.3} &\textbf{77.1} &\textbf{78.9} &\textbf{85.9} &\textbf{87.7} &\textbf{82.8} &\textbf{85.9} &\textbf{72.0} &\textbf{73.9} &\textbf{68.8} &\textbf{70.2} &\textbf{72.2} &\textbf{73.7} &58.4 & 59.2 &82.6 &84.8 &\textbf{78.6} &\textbf{81.5} &\textbf{65.0} &\textbf{68.8} &\textbf{83.0} &\textbf{84.8} &\textbf{74.0} &\textbf{76.1}\\
			\bottomrule
		\end{tabular}}
		\vspace{-2ex}
	\end{center}
\end{table*}

\subsection{Evaluation Metrics}

To evaluate the proposed method and the baselines, we utilize three widely used measures~\cite{OSBP, STA}, \textit{i.e.}, accuracy of the unknown class (\textbf{UNK}), normalized accuracy for all classes (\textbf{OS}), normalized accuracy for the known classes only (\textbf{OS$^*$}) and harmonic mean accuracy (\textbf{H}):
\begin{equation}
\begin{split}
     &\text{UNK} = \frac{|x: x \in \mathcal{D}^{C+1}_t \land  h(x)=C+1|}{|(x, y): (x, y) \in \mathcal{D}^{C+1}_t|},\\
    &\text{OS} = \frac{1}{C+1}\sum_{i=1}^{C+1}\frac{|x: x \in \mathcal{D}^i_t \land h(x)=i|}{|x: x \in \mathcal{D}^i_t|},\\
    &\text{OS}^* = \frac{1}{C}\sum_{i=1}^{C}\frac{|x: x \in \mathcal{D}^i_t \land h(x)=i|}{|x: x \in \mathcal{D}^i_t|},\\
    &\text{H} = \frac{2\times \text{OS}^* \times \text{UNK}}{\text{OS}^* + \text{UNK}}
\end{split}
\end{equation}
with $\mathcal{D}_t^i$ being the set of target samples in the $i$-th class, and $h(\cdot)$ the classifier. In our case, we use the shared classifier $\tilde{h}$ for the known classes and pseudo-labeling function $h_b$ for the unknown one. Notably, \textbf{H} is considered as the fairest evaluation metric, which trades off between the performance of the methods on known and unknown class samples.

In addition to classification accuracy, we explore the calibration capacity of the adapted model by leveraging the metric of expected calibration error (ECE). Given the model predictions and its respective confidence scores, ECE is calculated by grouping test data into M interval bins of equal size. Let $B_m$ be the set of indices whose maximum prediction score falls into the $m$-th bin. The accuracy and average confidence for $B_m$ are defined as,
\begin{equation}
    \begin{split}
        &\texttt{acc}(B_m) = \frac{1}{|B_m|} \sum_{i\in B_m} \mathfrak{1}(\tilde{y}_i = y_i),\\
        &\texttt{conf}(B_m) = \frac{1}{|B_m|} \sum_{i\in B_m} \hat{p}(y|x),
    \end{split}
\end{equation}
where $\hat{p}(y|x)$ is the highest confidence of the sample $i$. Given the accuracy and confidence scores for each bin, ECE is computed as the weighted sum of the mismatch over bins,
\begin{equation}
    \texttt{ECE} = \sum_{m=1}^M \frac{|B_m|}{N} |\texttt{acc}(B_m) - \texttt{conf}(B_m)|,\label{eq:ece}
\end{equation}
where $N$ is the total number of samples. 

\subsection{Implementation Details}
PyTorch implementation of the proposed PGL is available in a GitHub repository\footnote{https://github.com/BUserName/PGL} and the source code of SF-PGL is also made available\footnote{https://github.com/Luoyadan/SF-PGL}.  In our experiments, we employ ResNet-50, ResNet-152 \cite{resnet} and VGGNet \cite{VGG} pre-trained on ImageNet as the backbone network. For VGGNet, we only fine-tune the parameters in FC layers. The networks are trained with the ADAM optimizer with a weight decay of $5\times10^{-5}$. The learning rate is initialized as $1\times10^{-4}$ and $1\times10^{-5}$ for the GNNs and the backbone module, respectively, and then decayed by a factor of $0.5$ every $4$ epochs. The dropout rate is fixed to $0.2$ and the depth of GNN $L$ is set to $1$ for all experiments. The loss coefficients $\gamma$ and $\mu$ are empirically set to $0.4$ and $0.3$, respectively. \textcolor{black}{The threshold $\beta$ for PGL is set to $0.6$, $0.85$ and $0.9$ for the Office-Home, VisDA-17 and Syn2Real-O, respectively. The threshold $\beta$ for SF-PGL is set to $0.3$ and $0.9$ for the VisDA-17 and Syn2Real-O.} The batch sizes of the proposed PGL are set to 2, 8, 6 for three open-set benchmarks. The batch sizes of SF-PGL are fixed to 4 and 8 for VisDA-17 and Syn2Real-O datasets. The enlarging factor $\alpha$ is 0.05 The image feature extracted by the fc7 layer of VGGNet backbone is a 4096-D vector, and the deep feature extracted from the ResNet-50 is a 2048-D vector. For video tasks, we set the batch size to 12 for two UCF-HMDB datasets, UCF$\rightarrow$Olympic task, 10 for UCF$\rightarrow$Olympic task and 8 for Gameplay$\rightarrow$Kinetics. More details can be found in the Github repository for reproduction.

\subsection{Results of Domain-adaptive Image Classification}\label{sec:exp}
To validate the effectiveness of the proposed PGL and SF-PGL models, we compare them with state-of-the-art OSDA and SF-OSDA approaches. As reported in Table~\ref{tab:home}, Table~\ref{tab:visda17}, and Table~\ref{tab:Syn2Real-O}, we clearly observe that our method \textbf{PGL} consistently outperforms the state-of-the-art results, improving mean accuracy (OS$^*$) by $8.5\%$, $28.0\%$ and $28.0\%$ on the benchmark datasets of \textit{Office-Home}, \textit{VisDA-17} and \textit{Syn2Real-O} datasets respectively. Note that our proposed approach provides significant performance gains for the more challenging datasets of \textit{Syn2Real-O} and \textit{VisDA-17} which require knowledge transfer across different modalities. This phenomenon can be also observed in the transfer sub-tasks with a large domain shift \textit{e.g.}, \textbf{Rw}$\to$\textbf{Cl} and \textbf{Pr}$\to$\textbf{Ar} in \textit{Office-Home}, which demonstrates the strong adaptation ability of the proposed framework. For the SF-OSDA task, the proposed SF-PGL model surpasses not only the \textbf{Inherit} approach but also all OSDA methods by a large margin, as reported in the last row of Table \ref{tab:visda17} and row 5-7, row 11-13 of Table \ref{tab:Syn2Real-O}. Compared with the original PGL model, SF-PGL is capable of balancing the learning of `easy' and `difficult' concepts  by weighting the classification loss with the uncertainty-aware coefficient $w$. For instance, by comparing the results shown in row 4 and row 7 of Table \ref{tab:Syn2Real-O}, the mean accuracies of the Person and Knife categories are improved from $41.0\%$ to $94.8\%$ and from $45.4\%$ to $84.5\%$.

\begin{table}[t]
	\begin{center}
	\setlength{\tabcolsep}{3.7pt}     
    \setlength{\cmidrulekern}{0.25em} 
		\caption{Performance comparisons on the \textit{VisDA-17}. $^\dagger$ indicates methods with OSVM.}\label{tab:visda17}
		\resizebox{0.5\textwidth}{!}{
		\begin{tabular}{l c
            *{9}{S[table-format=2.1]} 
            }
			\toprule
			&\multicolumn{10}{c}{\textbf{VGG-based}}\\\cmidrule(lr){2-11}
			Method
			&Bic &Bus &Car &Mot &Tra &Tru &UNK &OS &OS$^*$ &H\\
			\midrule
			OSVM\cite{OSVM}  &31.7 &51.6 &66.5 &70.4 &88.5 &20.8  &38.1 &52.5 &54.9 &45.0\\
			MMD$^\dagger$\cite{MMD} &39.0 &50.1 &64.2 &79.9 &86.6 &16.3 &44.8 &54.4 &56.0 &49.8 \\
			DANN$^\dagger$\cite{DANN} &31.8 &56.6 &71.7 &77.4 &87.0 &22.3 &41.9 &55.5 &57.8 &48.6\\
			ATI-$\lambda^\dagger$\cite{DBLP:journals/pami/BustoIG20} &46.2 &57.5& 56.9& 79.1& 81.6& 32.7 &65.0 & 59.9 & 59.0 &61.9\\
			OSBP\cite{OSBP} & 51.1 & 67.1 & 42.8 & 84.2& 81.8 & 28.0 &85.1 & 62.9& 59.2 &69.8\\
			STA\cite{STA} & 52.4 & 69.6 & 59.9 & 87.8 & 86.5 & 27.2 &84.1 & 66.8 & 63.9 &72.6\\
			\midrule
			Inherit\cite{DBLP:conf/cvpr/KunduVRVB20} &53.5 &69.2 &62.2 &85.7 &85.4 &32.5	&\textbf{88.5} &68.1	&64.7 &74.8\\
			\textbf{PGL} &\textbf{93.5} &\textbf{93.8} &\textbf{75.7} &\textbf{98.8} &\textbf{96.2} &\textbf{38.5}&68.6 &\textbf{80.7} &\textbf{82.8} &\textbf{75.0}\\
			\midrule
			\midrule
            &\multicolumn{10}{c}{\textbf{ResNet50-based}}\\\cmidrule(lr){2-11}
            Method
			&Bic &Bus &Car &Mot &Tra &Tru &UNK &OS &OS$^*$ &H\\
			\midrule
            OSVM~\cite{OSVM}	&40.2	&55.4	&63.5	&70.8	&74.1	&35.2	&45.6	&54.9	&56.5	&50.5\\
            DANN$^\dagger$\cite{DANN}	&32.4	&51.6	&65.1	&71.3	&85.1	&23.1	&\text{ -}	& \text{ -}	&52.1	& \text{ -} \\
            RTN$^\dagger$\cite{RTN}	&31.6	&63.6	&54.2	&76.9	&87.3	&21.5	&\text{ -}	& \text{ -}	&51.1	& \text{ -}\\
            IWAN$^\dagger$\cite{IWAN}	&30.6	&69.8	&58.3	&76.8	&65.5	&30.8	&69.7	&57.3	&55.3	&61.7\\
            ETN$^\dagger$\cite{ETN}	&31.6	&66.8	&61.7	&77.8	&70.8	&30.8	&70.7	&58.6	&56.6	&62.9\\
            UAN\cite{DBLP:conf/cvpr/YouLCWJ19}	&42.6	&67.8	&65.7	&76.9	&69.8	&31.8	&70.7	&60.9	&59.1	&64.4 \\
            ATI-$\lambda^\dagger$\cite{DBLP:journals/pami/BustoIG20}	&33.6	&51.6	&64.2	&78.1	&85.3	&22.5	&42.5	&54.8	&52.6	&47.0\\
            OSBP\cite{OSBP}	&35.6	&59.8	&48.3	&76.8	&55.5	&29.8	&81.7	&55.4	&50.9	&62.7\\
            STA\cite{STA}	&50.1 &69.1	 &59.7	&85.7	&84.7	&25.1	&\textbf{82.4}	&65.3	&62.4	&71.0 \\
            \midrule
           \textbf{SF-PGL} w/o BP-L &91.5 &90.1	&74.1	&90.3	&81.9	&74.8	&72.0	&82.1	&83.8	&77.4\\
           \textbf{SF-PGL} &\textbf{93.6}	&\textbf{97.6} &\textbf{89.6}	&\textbf{95.3} &\textbf{96.7}	&\textbf{95.6} &68.6 &\textbf{91.0} &\textbf{94.7} &\textbf{79.6}\\
			\bottomrule
		\end{tabular}
		}
		\vspace{-3ex}
	\end{center}
\end{table}

 \begin{table}[t]
	\begin{center}
		\caption{Ablation performance on the \textit{Syn2Real-O} (ResNet-50). ``w'' indicates with and ``w/o'' indicates without. }\label{tab:abl} 
		\resizebox{1\linewidth}{!}{
		\begin{tabular}{l cc cc}
			\toprule
    		Model
			&UNK  & OS& OS$^*$ &H-Score\\
			\midrule
			PGL w/o Progressive &43.6 &54.4 & 55.3 &48.8\\
			PGL w NLL & 48.6  &56.9 &57.6 &52.7\\
			PGL w/o GNNs & 49.2& 57.8 &58.5 &53.4\\
			PGL w/o Mix-up &\textbf{49.8}&62.5&63.6 &55.9\\
			\midrule
			\midrule
		    \textbf{PGL}  &49.6 &\textbf{65.5} &\textbf{66.8} &\textbf{56.9}\\
			\bottomrule
		\end{tabular}}
	\end{center}
\end{table}
\begin{table*}[t]
	\begin{center}
		\caption{Recognition accuracies (\%) for open-set domain adaptation experiments on the \textit{Syn2Real-O} (ResNet-50). }\label{tab:Syn2Real-O}
		\setlength{\tabcolsep}{3.7pt}     
        \setlength{\cmidrulekern}{0.25em} 
		\resizebox{0.92\textwidth}{!}{
		\begin{tabular}{l c c c c c c c c c c c c c c c c}
			\toprule
            &\multicolumn{16}{c}{ResNet50-based}\\\cmidrule(lr){2-17}
			Method&Aer&
			Bic&
			Bus&
			Car&
			Hor&
			Kni&
			Mot&
			Per&
			Pla&
			Ska&
			Tra&
			Tru&
			UNK&
			OS&
			OS$^*$ &H-Score\\
			\midrule
			DANN$^\dagger$~\cite{DANN} & 50.8&44.1&19.0&58.5&76.8 &26.6&68.7 &\textbf{50.5}&82.4&21.1&69.7&1.1&33.6&46.3&47.4 &39.3\\
			OSBP~\cite{OSBP} &75.5 &67.7 &68.4 &\textbf{66.2}&71.4&0.0&86.0&3.2&39.4&23.2&68.1&3.7&\textbf{79.3}&50.1&47.7 &59.6\\
			STA~\cite{STA} & 64.1& \textbf{70.3}& 53.7& 59.4 & 80.8& 20.8 & 90.0& 12.5& 63.2& 30.2& 78.2& 2.7&59.1&52.7&52.2 &55.4\\
			\midrule
			\textbf{PGL} &81.5 &68.3 &74.2 & 60.6 &91.9 &45.4 &92.2 &41.0 &87.9 &67.5 &79.2 &\textbf{6.4} &49.6 &65.5 &66.8 &56.9\\
			\textbf{SF-PGL} ($\alpha=0.2$) &82.7	&84.7	&86.1	&82.0	&87.9	&49.8	&82.7	&86.8	&\textbf{88.8}	&67.1	&77.3	&0.0	&85.0	&73.9	&73.0	&78.5\\
			\textbf{SF-PGL} ($\alpha=0.1$) &\textbf{91.5}	&88.8	&90.4	&87.4	&89.8	&80.7	&92.2	&88.5	&74.9	&87.3	&87.2	&0.0	&88.9	&80.6	&79.9	&84.2\\
            \textbf{SF-PGL} ($\alpha=0.05$)	&90.7	&\textbf{90.5}	&\textbf{93.0}	&\textbf{90.0}	&\textbf{93.6}	&\textbf{84.5}	&\textbf{92.6}	&\textbf{94.8}	&\textbf{88.8}	&\textbf{93.4}	&\textbf{91.7}	&0.0	&\textbf{94.3}	&\textbf{84.5}	&\textbf{83.6}	&\textbf{88.6}\\
            \midrule
            \midrule
            &\multicolumn{16}{c}{ResNet152-based}\\\cmidrule(lr){2-17}
			Method&Aer&
			Bic&
			Bus&
			Car&
			Hor&
			Kni&
			Mot&
			Per&
			Pla&
			Ska&
			Tra&
			Tru&
			UNK&
			OS&
			OS$^*$ &H-Score\\
			\midrule
			OSVM~\cite{OSVM} &53.8	&54.2	&50.3	&48.7	&72.7	&5.3	&82.0	&27.0	&49.6	&43.4	&78.0	&\textbf{5.1}	&44.2	&47.3	&47.5	&45.8\\
			OSBP~\cite{OSBP} &80.2	&63.1	&59.1	&63.1	&83.2	&12.1	&89.1	&5.0	&61.0	&14.0	&79.2	&0.0	&69.0	&52.2 &50.8	&58.5\\
			SE-CC~\cite{DBLP:conf/cvpr/PanYLNM20} &82.1	&80.7	&59.7	&50.0	&80.6	&36.7	&83.1	&56.2	&56.6	&21.9	&57.7	&4.0	&70.6	&56.9	&55.8	&62.3\\
			\midrule
			\textbf{SF-PGL} ($\alpha=0.2$)	&94.8	&87.9	&90.0	&83.5	&84.0	&49.3	&92.2	&82.8	&76.9	&91.2	&93.1	&0.2	&84.2	&77.7	&77.2	&80.5\\
            \textbf{SF-PGL} ($\alpha=0.1$)	&97.5	&90.5	&94.7	&89.6	&90.0	&76.8	&\textbf{95.0}	&91.3	&84.1	&95.2	&96.9	&0.0	&85.3	&83.6	&83.5	&84.4\\
            \textbf{SF-PGL} ($\alpha=0.05$)	&\textbf{98.4}	&\textbf{95.4}	&\textbf{96.4}	&\textbf{93.9}	&\textbf{93.0}	&\textbf{88.4}	&\textbf{95.0}	&\textbf{93.2}	&\textbf{87.2}	&\textbf{98.2}	&\textbf{98.3}	&0.0	&\textbf{85.0}	&\textbf{86.4}	&\textbf{86.5}	&\textbf{86.4}\\
    		\bottomrule
		\end{tabular}
		}
	\end{center}
\end{table*}

\begin{figure}
	\centering
	\includegraphics[width=1\linewidth]{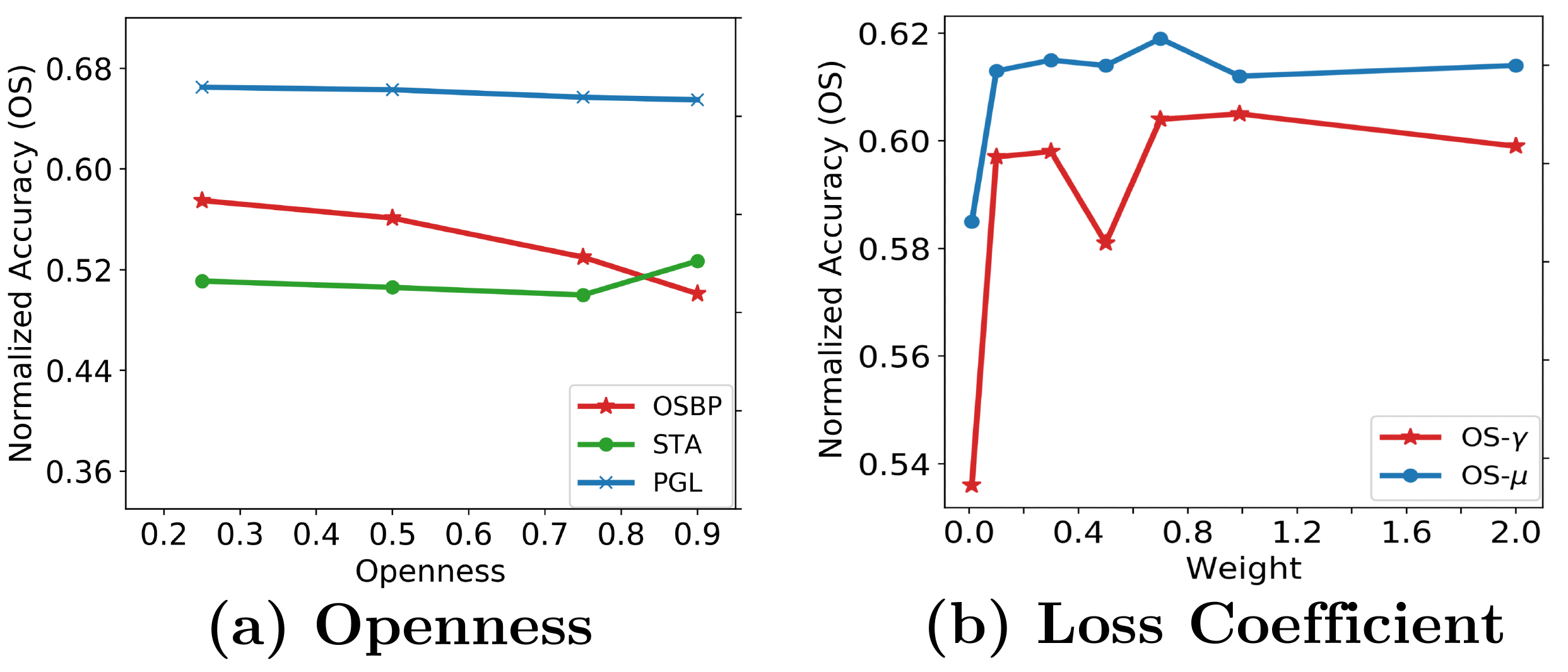}
	\caption{Performance Comparisons \textit{w.r.t.} varying (a) openness of the \textit{Syn2Real-o} (ResNet-50); (b) loss coefficients $\mu$ and $\gamma$ on the Ar$\to$Cl task (\textit{Office-Home}) with the ResNet-50 backbone.} \label{fig:analysis}
	\vspace{-1ex}
\end{figure}

\subsection{Model Analysis of PGL}
\textbf{Ablation Study:}
To investigate the impact of the derived progressive paradigm, GNNs, node classification loss, and mix-up strategy, we compare four variants of the PGL model on the \textit{Syn2Real-O} dataset shown in Table \ref{tab:abl}. Except for \textbf{PGL w/o Progressive} that takes $\alpha = 1$ and $\beta=0.6$, all experiments are conducted under the default setting of hyperparameters. \textbf{PGL w/o Progressive} corresponds to the model directly trained with one step, followed by pseudo-labeling function for classifying the unknown samples. \textcolor{black}{The results of \textbf{PGL w/o Progressive} on other datasets (\textit{i.e.}, \textit{Office-Home} and \textit{Syn2Real}) can be found in our Github repository.} As shown in Table \ref{tab:abl}, without applying the progressive learning strategy, the OS result of \textbf{PGL w/o Progressive} significantly drops by 16.9\% because \textbf{PGL w/o Progressive} does not leverage the pseudo-labeled target samples leading to the failure in minimizing the shared error at the sample-level. In \textbf{PGL w NLL}, the focal loss of the node classification objective is replaced with the Negative log-likelihood (NLL) loss, resulting in OS performance dropping from 65.5\% to 56.9\%. Due to the absence of the focal loss re-weighting module, the model tends to assign more pseudo-labels to easy-to-classify samples, which consequently hinders effective graph learning in the episodic training process. In \textbf{PGL w/o GNNs}, we used ResNet-50 as the backbone for feature learning, which triggers 12.5\% OS performance drops compared to the graph learning model. The inferior results reveal that the GNN module can learn the class-wise manifold, which mitigates the potential noise and permutation by aggregating the neighboring information. \textbf{PGL w/o Mix-up} refers to the model that constructs episodes without taking any pseudo-labeled target data. We observe that the OS performance of \textbf{PGL w/o Mix-up} is 4.6\% lower than the proposed model, confirming that replacing the source samples with pseudo-labeled target samples progressively can alleviate the side effect of conditional shift.

\begin{table}
	\begin{center}
		\caption{Performance comparisons \textit{w.r.t.} varying enlarge factor $\alpha$ on the \textit{Syn2Real-O} and \textit{Office-Home} (ResNet-50).}\label{tab:EF} \vspace{1ex}
		\resizebox{1\linewidth}{!}{
		\begin{tabular}{c cc cc}
			\toprule
			\multirow{2}{*}{\textbf{Enlarging Factor}}&
			\multicolumn{2}{c}{~~~~\textbf{Syn2Real-O}~~~~}&
			\multicolumn{2}{c}{\textbf{Office-Home (Ar-Cl)}}\\
			\cmidrule(l){2-3}
			\cmidrule(l){4-5}
			&~OS~~& OS$^*$& ~~OS~~~~~& OS$^*$\\
			\midrule
			$\alpha = 0.20$ & 63.0 & 63.3 & 59.9 & 61.1\\
			$\alpha  = 0.10$ & 64.5 & 65.7 & 60.7 & 61.6\\
		    $\alpha  = 0.05$ &\textbf{65.6} &\textbf{66.5} & \textbf{61.8} & \textbf{63.1}\\
			\bottomrule
		\end{tabular}}
	\end{center}
\end{table}

\begin{figure}[t]
	\centering
	{\includegraphics[width=.92\linewidth]{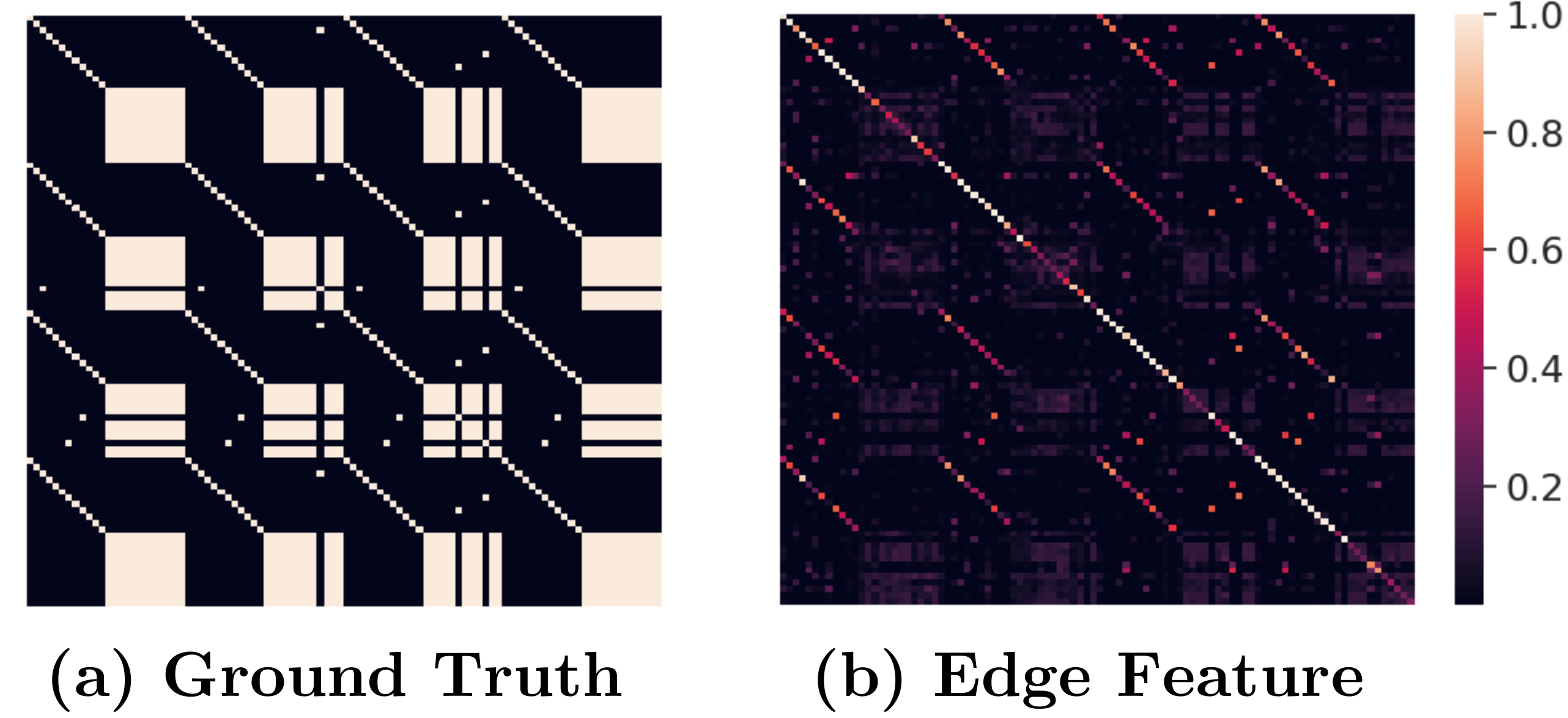}}
	\caption{Visualization of edge features on the \textit{Syn2Real-O}. \textit{Left}: the binary ground-truths label map. \textit{Right}: the learned edge map from the proposed edge update networks.  Best viewed in color.} \label{fig:edge}
	\vspace{-2ex}
\end{figure}
\begin{figure*}
    \centering
    \includegraphics[width=0.4\linewidth]{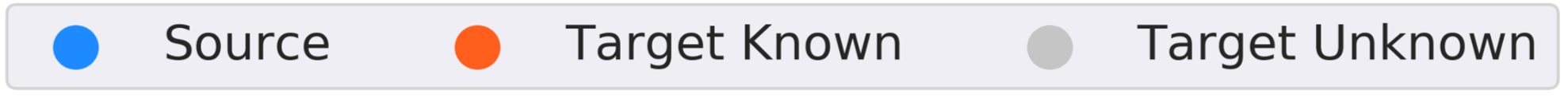}\\
    \subfloat[][ResNet-50]{\includegraphics[width=0.17\linewidth]{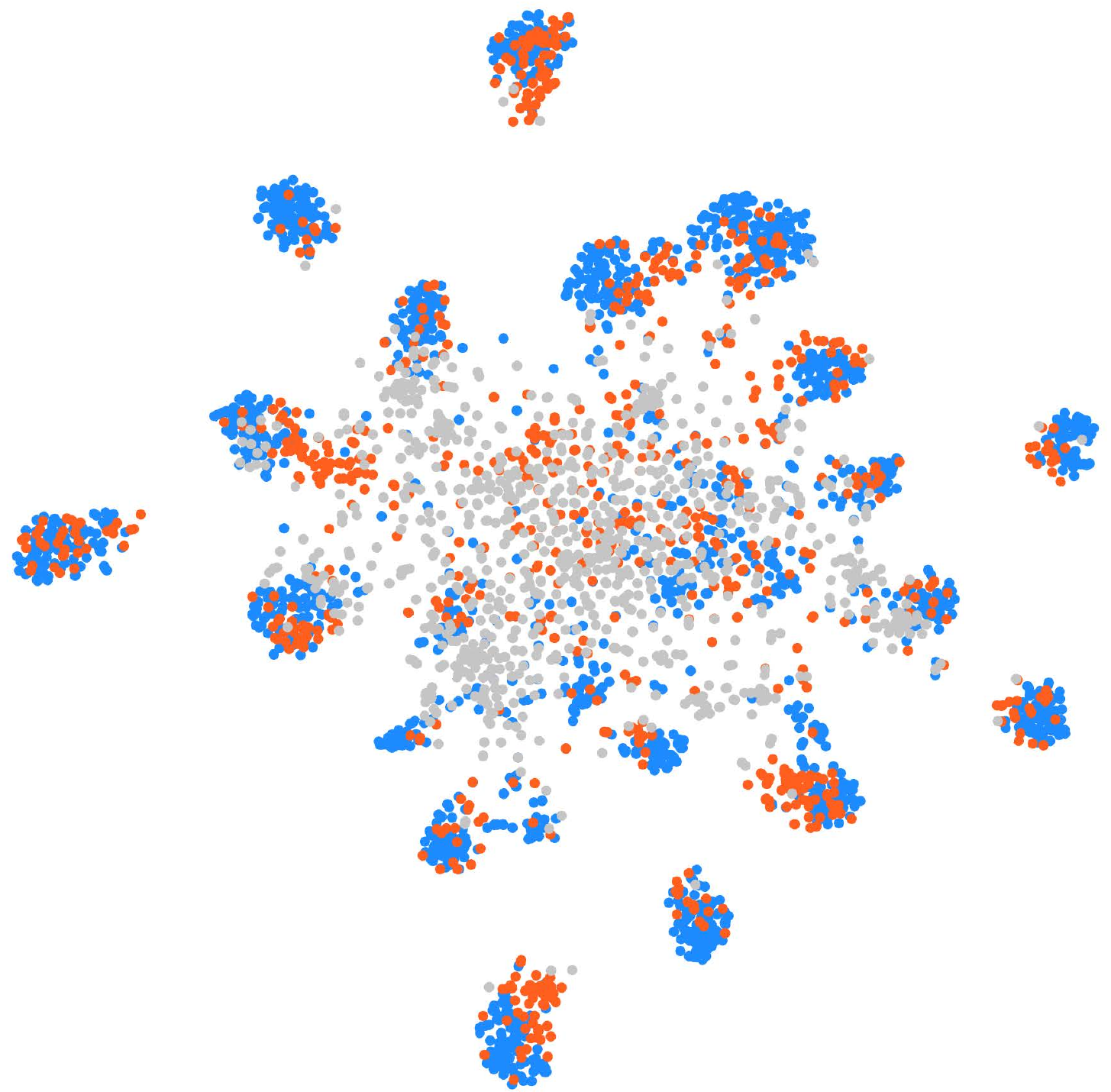}}\hspace{3ex}
    \subfloat[][DANN]{\includegraphics[width=0.17\linewidth]{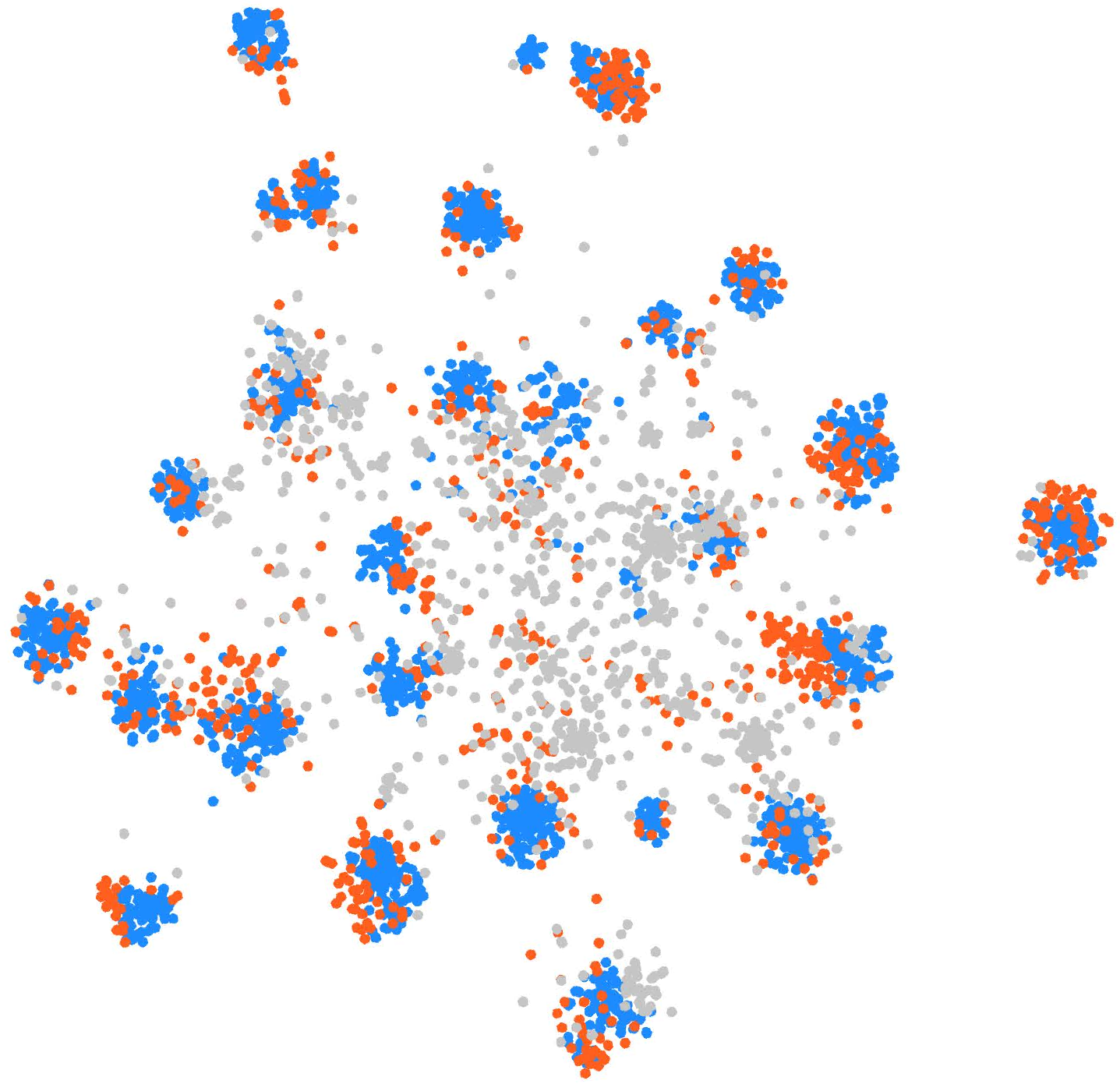}}\hspace{3ex}
    \subfloat[][OSBP]{\includegraphics[width=0.17\linewidth]{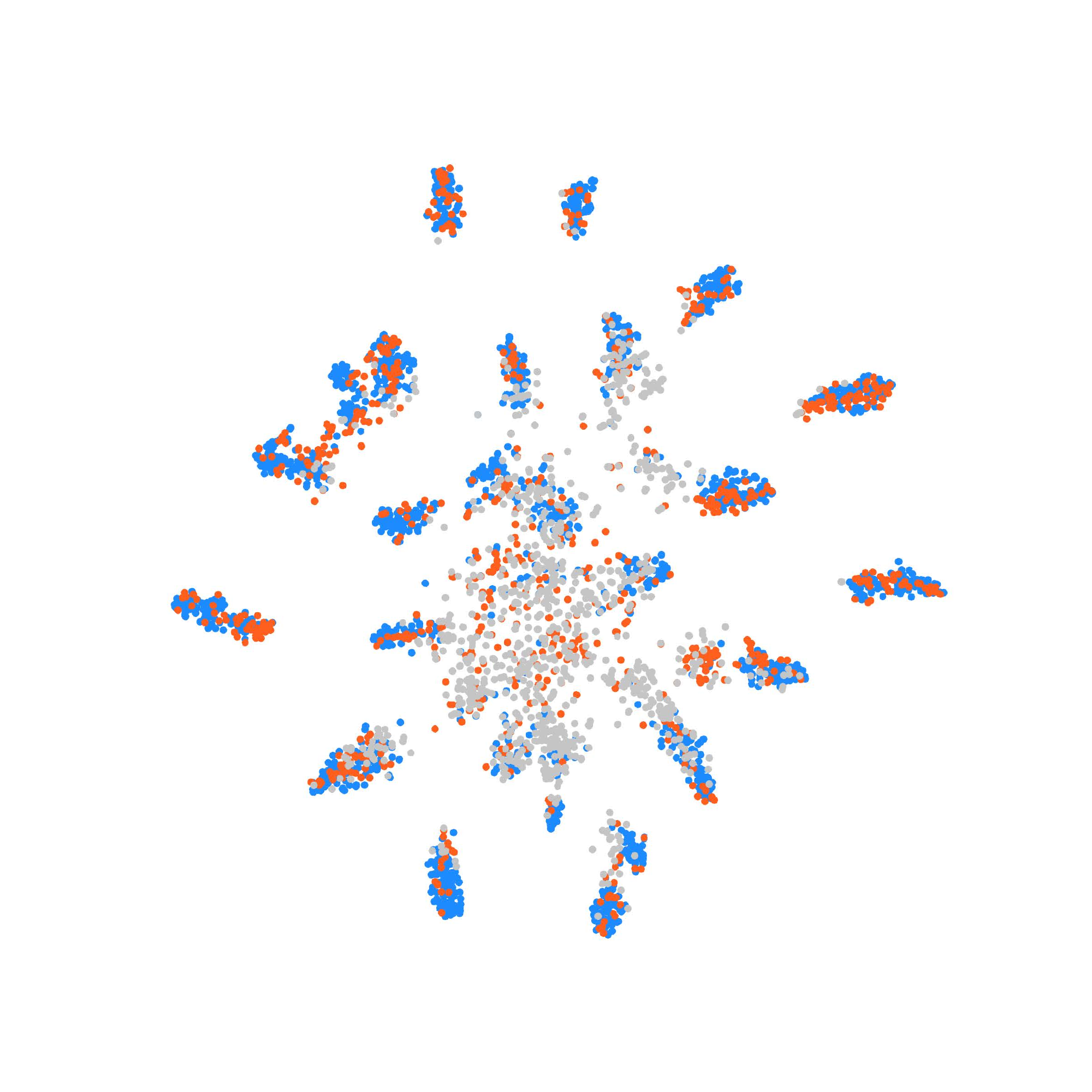}}\hspace{3ex}
    \subfloat[][STA]{\includegraphics[width=0.17\linewidth]{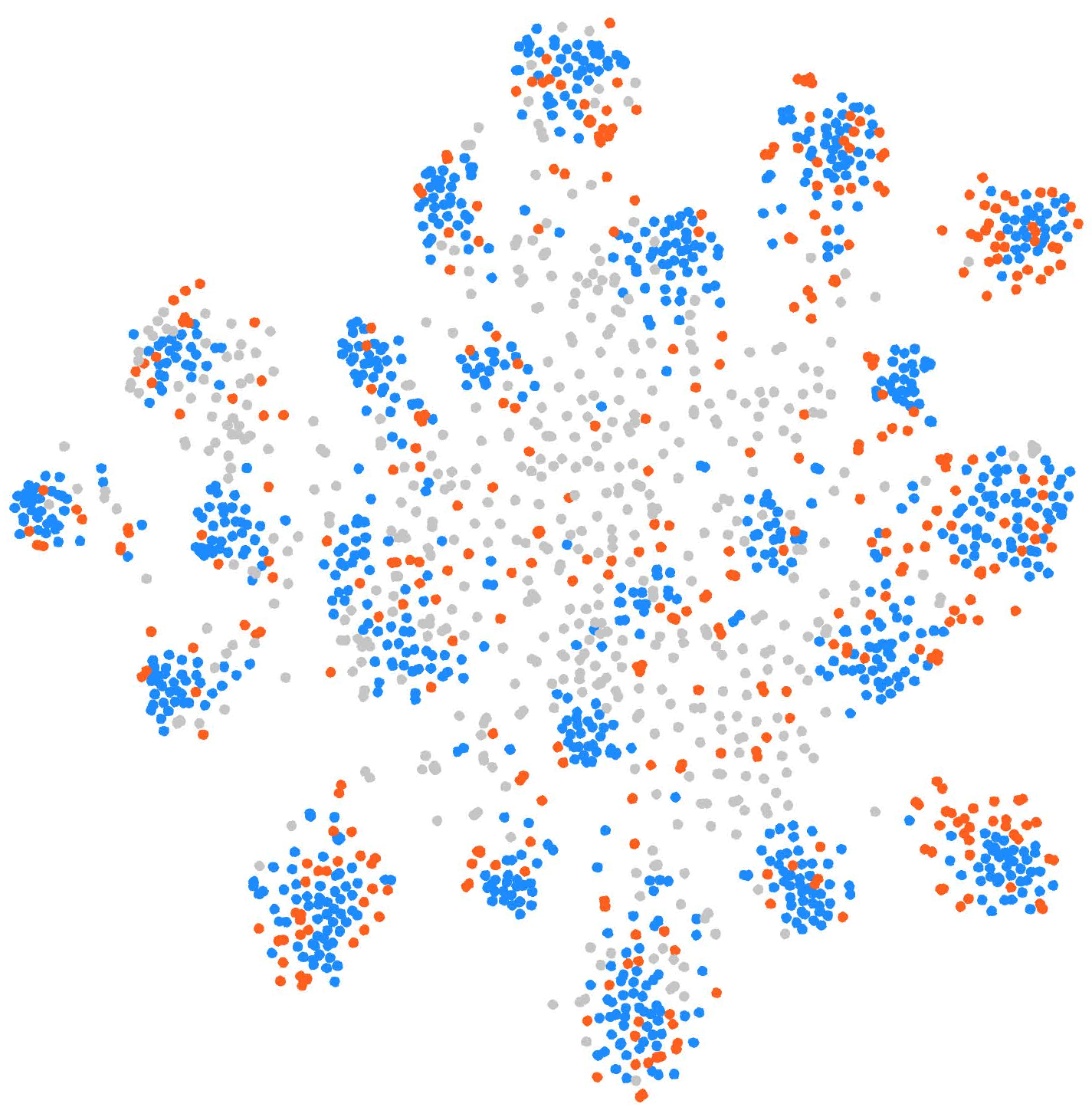}}\hspace{3ex}
    \subfloat[][PGL]{\includegraphics[width=0.17\linewidth]{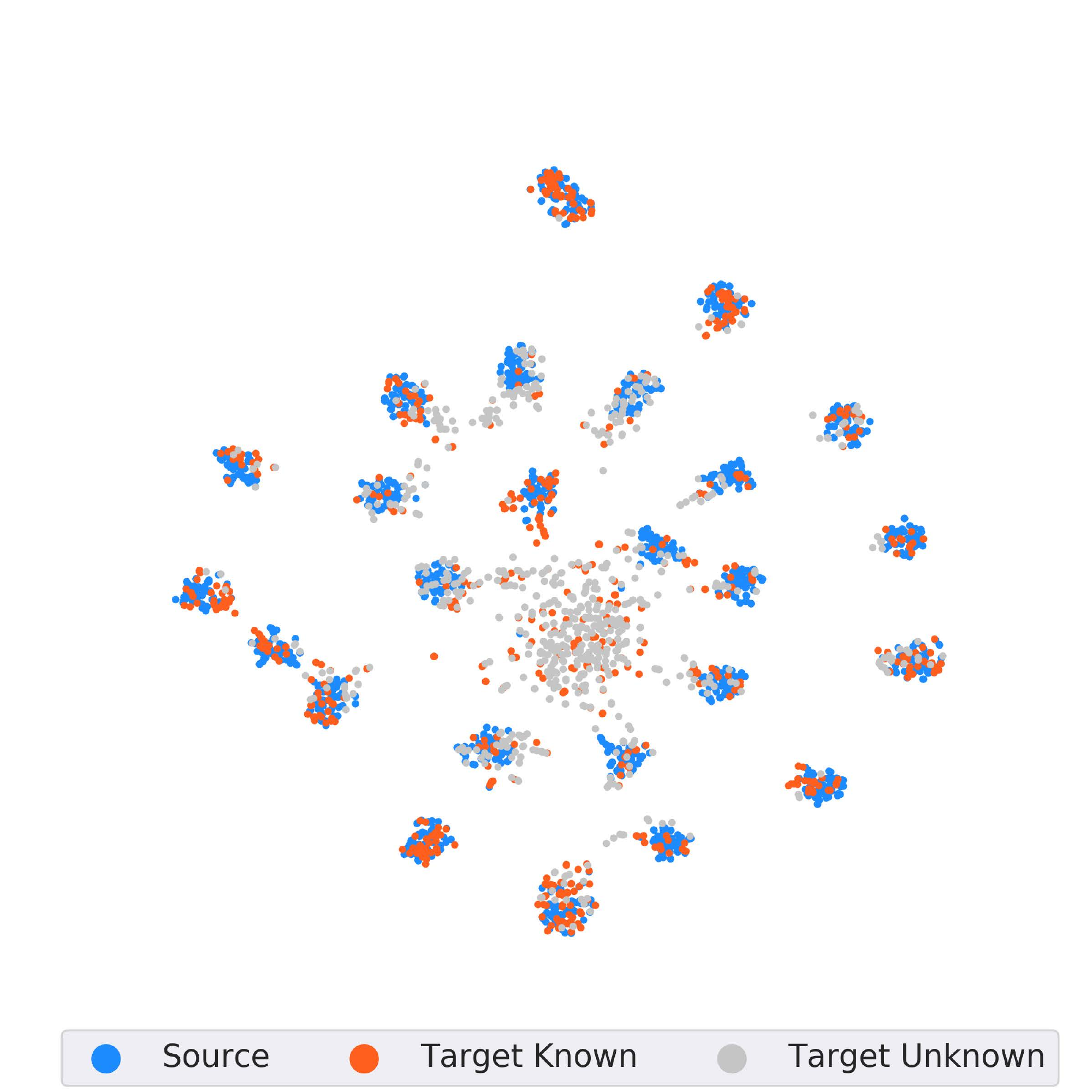}}
    \caption{The t-SNE visualization of feature distributions on the Rw$\to$Ar task (\textit{Office-Home}) with the ResNet-50 backbone.}
    \label{fig:tsne-home}
\end{figure*}

\textbf{Robustness Analysis to Varying Openness:}
To verify the robustness of the proposed PGL, we conduct experiments on the \textit{Syn2Real-O} with the openness varying in $\{0.25, 0.5, 0.75, 0.9\}$. The openness is defined as the ratio of unknown samples to all samples in the entire target set, which explicitly implies the level of challenge. The results of \textbf{OSBP}, \textbf{STA} and the proposed \textbf{PGL} are depicted in Fig. \ref{fig:analysis}(a). Note that \textbf{OSBP} and our \textbf{PGL} approach empirically sets a hyperparameter ($\beta$ in our case) to control the openness, while \textbf{STA} automatically generates the soft weight in adversarial way and inevitably results in performance fluctuation. We observe that \textbf{PGL} consistently outperforms the counterparts by a large margin, which confirms its resistance to the change in openness.



\textbf{Sensitivity to Loss Coefficients $\mu$ and $\gamma$.}
We show the sensitivity of our approach to varying the edge loss coefficient $\mu$ and adversarial loss coefficient $\gamma$ in Fig. \ref{fig:analysis}(b). We vary the value of one loss coefficient from (0, 2] at each time, while fixing the other parameter to the default setting. Two observations can be drawn from Fig. \ref{fig:analysis}(b): The OS score becomes stable when loss coefficients are within the interval of [0.7, 2]; When $\mu\rightarrow 0$, $\gamma\rightarrow 0$, the model performance drops by $4.6\%$ and $10.2\%$ respectively, which verifies the importance of the edge supervision and adversarial learning in our framework.


\textbf{Sensitivity to Enlarging Factor $\alpha$.}
We further study the effectiveness of the enlarging factor $\alpha$, which controls the enlarging speed of the pseudo-labeled set, shown in Table \ref{tab:EF}. We note that the proposed model with a smaller value of $\alpha$ consistently performs better on both the \textit{Syn2Real-O} and \textit{Office-Home} datasets. This testifies our theoretical findings that the progressive open-set risk $\tilde{\Delta}_o$ can be controlled by consecutively classifying unknown samples. With a sacrifice on the training time, this strategy also provides more reliable pseudo-labeled candidates for the shared classifier learning, preventing the potential error accumulation in the next several steps.

\textbf{t-SNE Visualization.} To intuitively showcase the effectiveness of OUDA approaches, we extract features from the baseline models (\textbf{ResNet-50}, \textbf{DANN}, \textbf{OSBP}, \textbf{STA}) and our proposed model \textbf{PGL}
on the Rw$\to$Ar task (\textit{Office-Home}) with the ResNet-50 backbone. The feature distributions are visualized with t-SNE afterward. As shown in Fig. \ref{fig:tsne-home}, compared with \textbf{ResNet-50} and \textbf{DANN}, open-set domain adaptation methods generally have a better separation between the known (in blue and red) and unknown (in grey) categories. \textbf{STA} achieves a better alignment between the source and target distributions in comparison with \textbf{OSBP}, while the \textbf{PGL} can obtain a clearer class-wise classification boundary, benefiting from our graph neural networks and the mix-up strategy. 

\begin{table}[t]\caption{Classification accuracies (\%) of the proposed PGL method for open-set action recognition. $^\dagger$ indicates methods with OSVM.}\label{tab:video}
\resizebox{1.01\linewidth}{!}{
\begin{tabular}{lcccc cccc}
\toprule
            & \multicolumn{4}{c}{HMDB$\rightarrow$UCF$_{small}$} & \multicolumn{4}{c}{UCF$\rightarrow$HMDB$_{small}$} \\ 
            \cmidrule(lr){2-5} \cmidrule(lr){6-9}
            & OS     & OS*    & UNK     & H  & OS     & OS*    & UNK     & H  \\
            \midrule
TA$^3$N$^\dagger$\cite{DBLP:conf/iccv/ChenKAYCZ19}        & 89.1   & 87.1   & 97.1    & 91.9     & 85.3   & 83.3   & 93.3    & 88.1     \\
OSBP\cite{OSBP}        & 90.1   & 87.6   & 100.0   & 93.4     & 89.9   & 87.4   & 100.0   & 93.3     \\
PGL w/o adv      & 87.2   & 95.4   & 94.3    & 89.6     & 91.8   & 89.7   & 100.0   & 94.6     \\
\textbf{PGL}         & \textbf{94.5}   & \textbf{94.5}   & \textbf{94.3}    & \textbf{94.4}     & \textbf{95.3}   & \textbf{94.9}   & \textbf{96.7}    & \textbf{95.8}     \\
\midrule
\midrule
            & \multicolumn{4}{c}{Olympic$\rightarrow$UCF}    & \multicolumn{4}{c}{UCF$\rightarrow$Olympic}    \\
            \cmidrule(lr){2-5} \cmidrule(lr){6-9}
            & OS     & OS*    & UNK     & H  & OS     & OS*    & UNK     & H  \\
            \midrule
TA$^3$N$^\dagger$\cite{DBLP:conf/iccv/ChenKAYCZ19}        & 73.7   & 72.9   & 77.6    & 75.1     & 81.7   & 80.9   & 85.7    & 83.2     \\
OSBP\cite{OSBP}        & 76.3   & 72.1   & 97.6    & 82.9     & 75.2   & 70.3   & 100.0   & 82.6     \\
PGL w/o adv	&87.2	&88.3	&81.6 &84.9 	&85.9	&83.1	&100.0	&90.7\\
\textbf{PGL}         &\textbf{89.8}  & \textbf{87.7}   & \textbf{100.0}   & \textbf{93.5}     & \textbf{95.3}   & \textbf{93.2}   &\textbf{ 100.0 }  & \textbf{96.5}     \\
\midrule
\midrule
            & \multicolumn{4}{c}{HMDB$\rightarrow$UCF$_{full}$}  & \multicolumn{4}{c}{UCF$\rightarrow$HMDB$_{full}$}  \\
            \cmidrule(lr){2-5} \cmidrule(lr){6-9}
            & OS     & OS*    & UNK     & H  & OS     & OS*    & UNK     & H  \\
            \midrule
DANN$^\dagger$\cite{DANN}        & 64.6   & 62.9   & 74.7    & 68.3     & 53.4   & 48.3   & \textbf{83.9}    & 61.3     \\
JAN$^\dagger$\cite{JAN}         & 64.5   & 62.9   & 73.8    & 67.9     & 51.6   & 47.8   & 74.4    & 58.2     \\
AdaBN$^\dagger$\cite{AdaBN}       & 60.9   & 58.8   & 73.4    & 65.3     & 61.4   & 61.7   & 59.4    & 60.5     \\
MCD$^\dagger$\cite{MCD}         & 65.0   & 63.5   & 73.8    & 68.3     & 60.3   & 57.8   & 75.6    & 65.5     \\
TA$^3$N$^\dagger$\cite{DBLP:conf/iccv/ChenKAYCZ19}        & 60.6   & 61.8   & 58.4    & 82.5     & 56.0   & 53.3   & 71.7    & 61.2     \\
OSBP\cite{OSBP}        & 62.3   & 58.4   & 85.9    & 69.5     & 56.3   & 51.3   & 78.9    & 64.4     \\
PGL w/o adv & 70.0   & 68.5   & 79.5    & 73.6     & 62.8   & 62.2   & 66.1    & 64.1     \\
\textbf{PGL}         & \textbf{81.3}   &\textbf{79.8}   & \textbf{90.8}    &\textbf{84.9}    & \textbf{69.8}   &\textbf{69.4}   & 71.7    & \textbf{70.5}    \\ \bottomrule
\end{tabular}
}
\end{table}

\begin{figure}[t]
	\centering
	\subfloat[][Office-Home]
	{\includegraphics[width=0.5\linewidth]{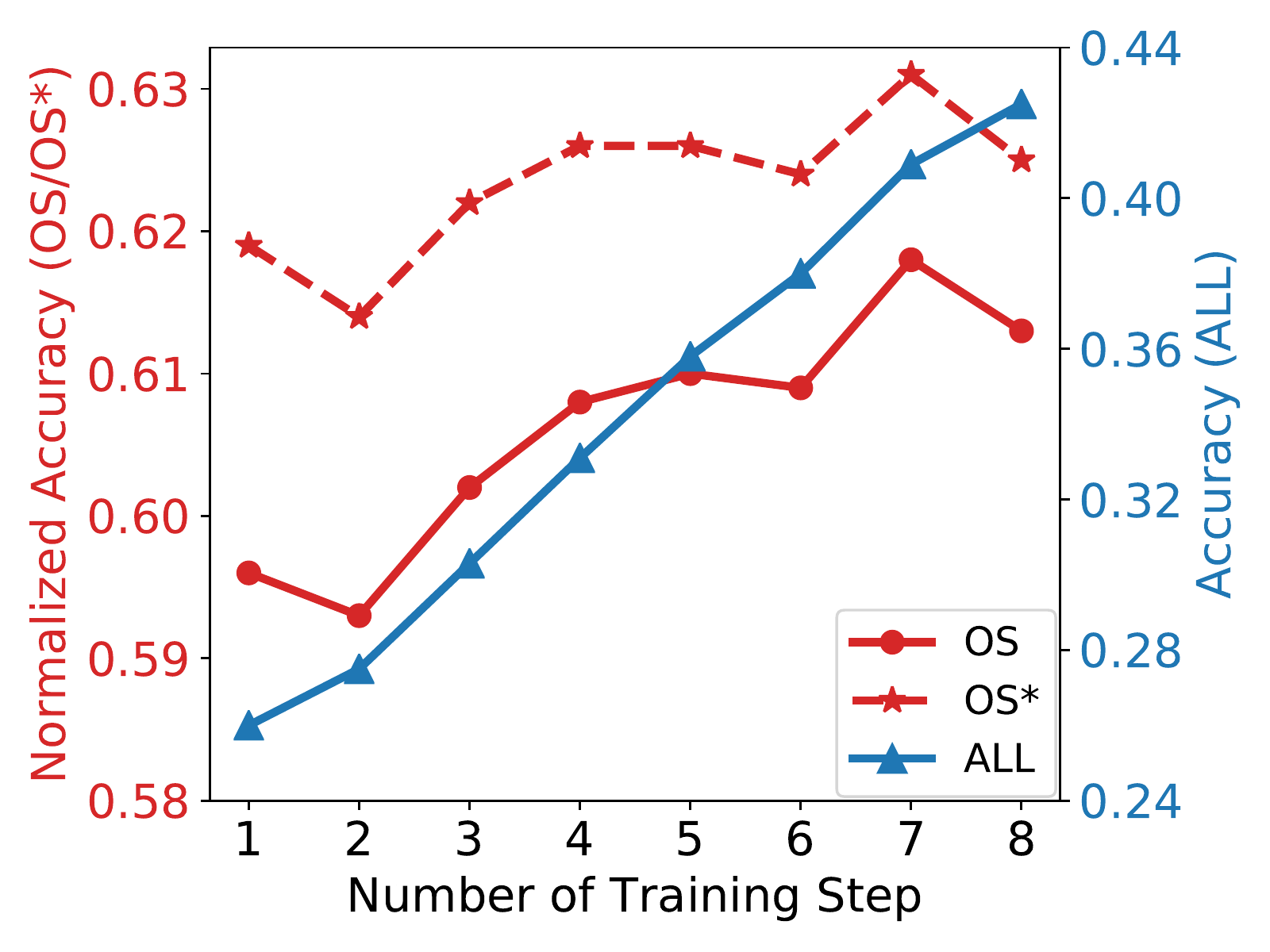}}
	\subfloat[][Syn2Real-O]
	{\includegraphics[width=0.5\linewidth]{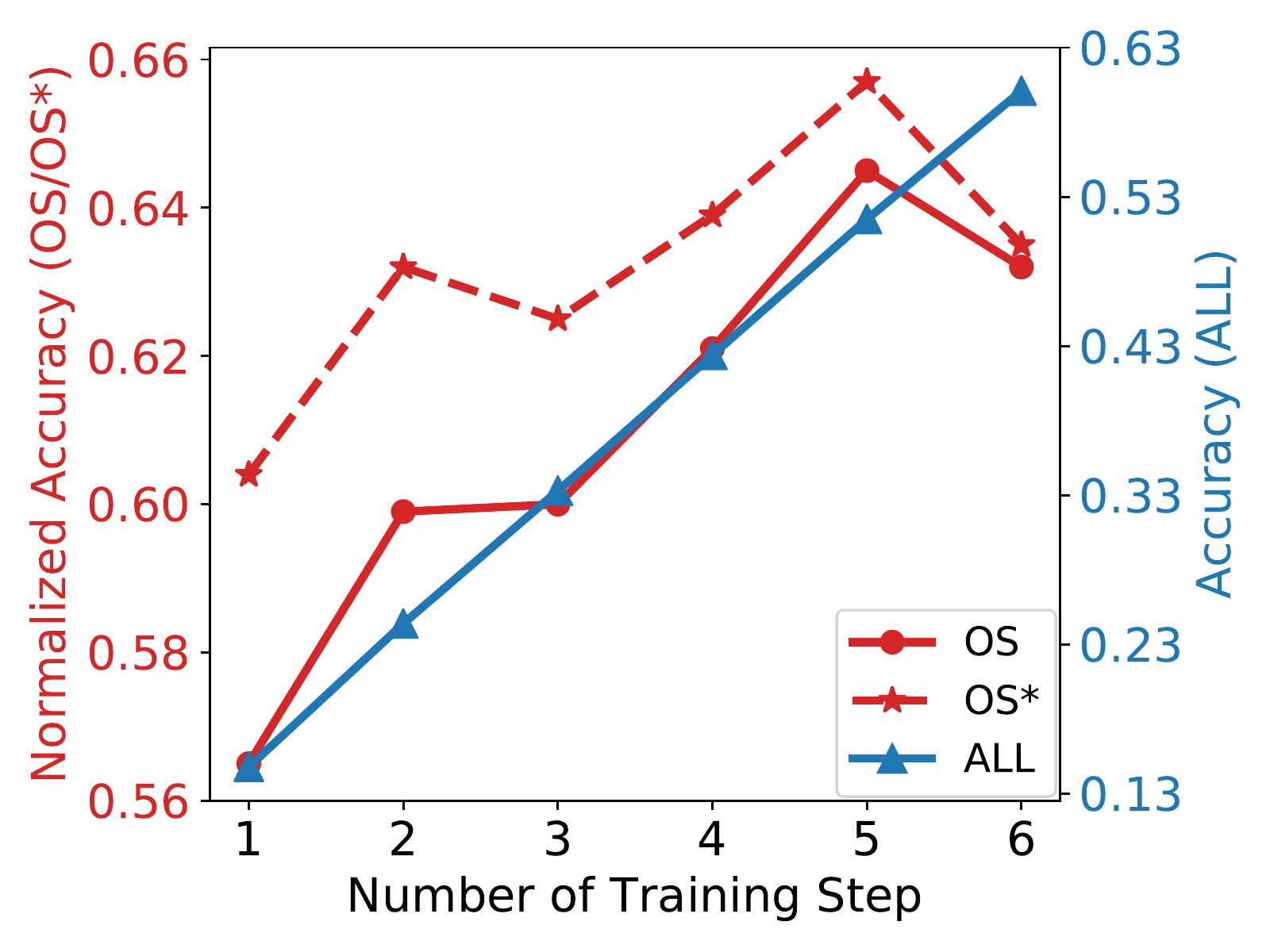}}
	\caption{Recognition accuracies of the proposed PGL method on the (a) Ar$\to$Cl task (\textit{Office-Home}) and (b) \textit{Syn2Real-O} datasets.} \label{fig:ep_analysis}
\end{figure}

  \textbf{Edgemap Visualization.} To further analyze the validity of the edge update networks, we extract the learned feature map  from the PGL with a single-layer GNN on the \textit{Syn2Real-O} dataset. As visualized in Fig. \ref{fig:edge}(b), a large value of $\mathcal{E}_{ij}$ corresponds to a high degree of correlations between node $v_i$ and $v_j$, which resembles the pattern of the ground-truth edge label $\widehat{Y}$ as displayed in Fig. \ref{fig:edge}(a).

\textbf{Quantitative Analysis over Training Steps.} Fig. \ref{fig:ep_analysis} illustrates the recognition performance of PGL over training steps on  the Ar$\to$Cl task of the \textit{Office-Home} dataset and \textit{Syn2Real-O} dataset, respectively. Three evaluation metrics are used to testify performance, \textit{i.e.}, the overall accuracy \textbf{ALL}, and normalized accuracies  \textbf{OS} and \textbf{OS$^*$}. All metrics gain a performance boost over the first several steps, as the pseudo-labeled target samples added in the source episodes can assist the classifier to make a more accurate prediction. Then, the normalized accuracy OS and OS$^*$ experience a downward because the enlarging pseudo-labeled set brings along noise and disturbance, which may degrade the model performance. In contrast, the accuracy ALL continuously increases as more unknown target samples are correctly classified, which occupy a large portion of the target domain. The results characterize a trade-off between normalized accuracy OS / OS$^*$ and the accuracy for unknowns. Considering the core value of domain adaptation is to correctly classify the classes of interest rather than irrelevant classes, we choose to stop the model updates at the training step 7 for the \textit{Office-Home} and the step 5 for the \textit{Syn2Real-O}.
\begin{figure*}[t]
    \centering
    \includegraphics[width=0.4\linewidth]{Images/TSNE/legend.png}\\
    \subfloat[][ResNet-50]{\includegraphics[width=0.15\linewidth]{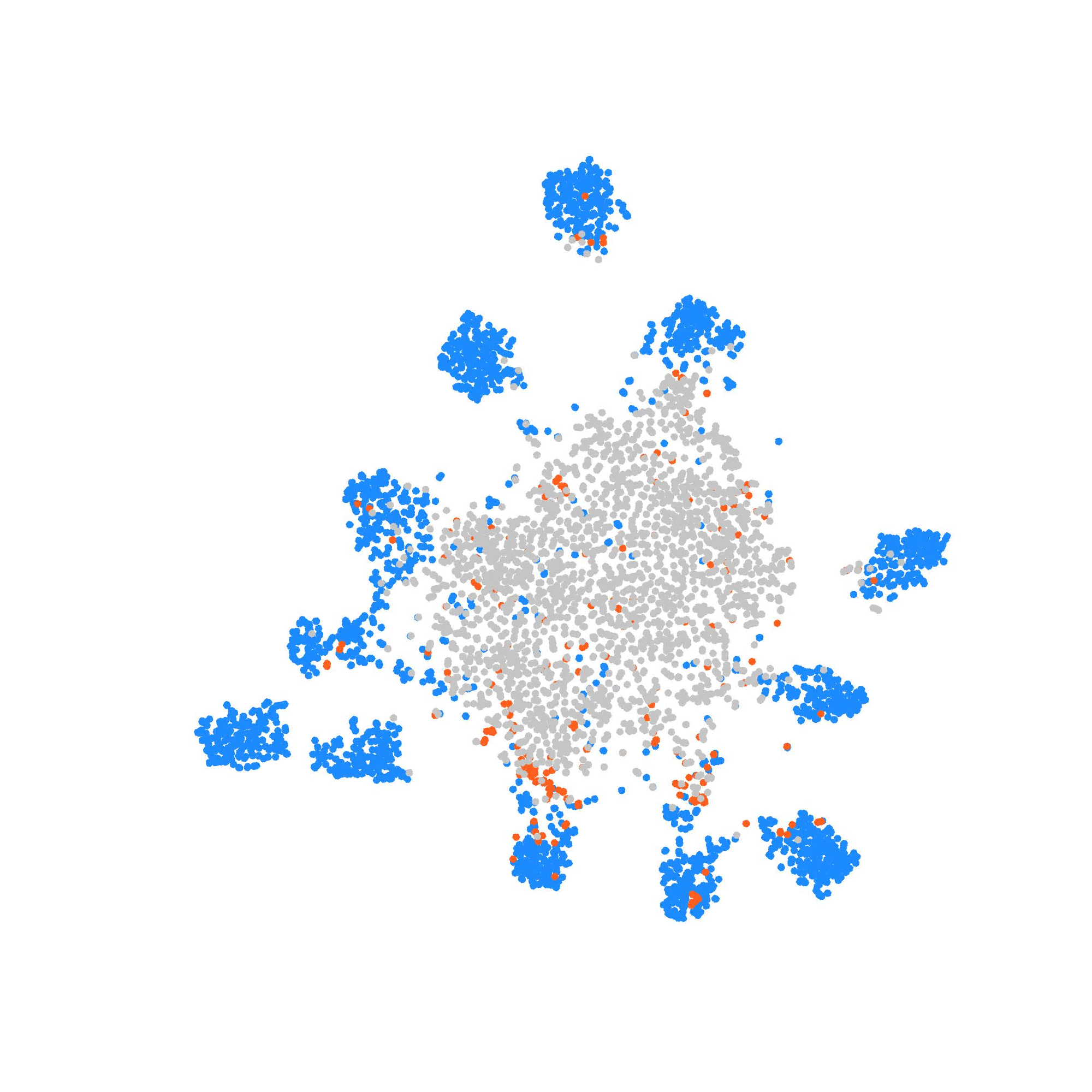}}\hspace{1ex}
    \subfloat[][DANN]{\includegraphics[width=0.15\linewidth]{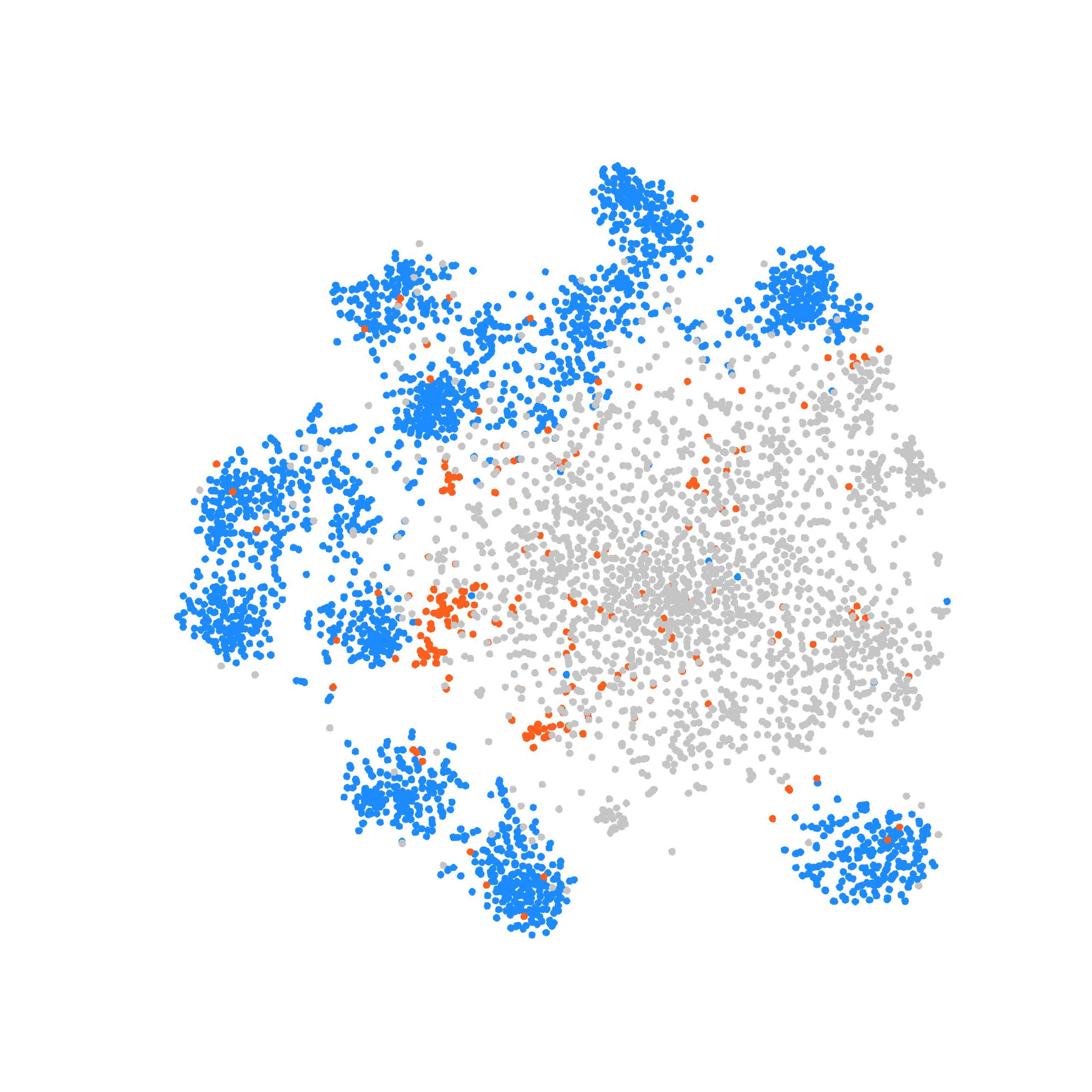}}\hspace{1ex}
    \subfloat[][OSBP]{\includegraphics[width=0.15\linewidth]{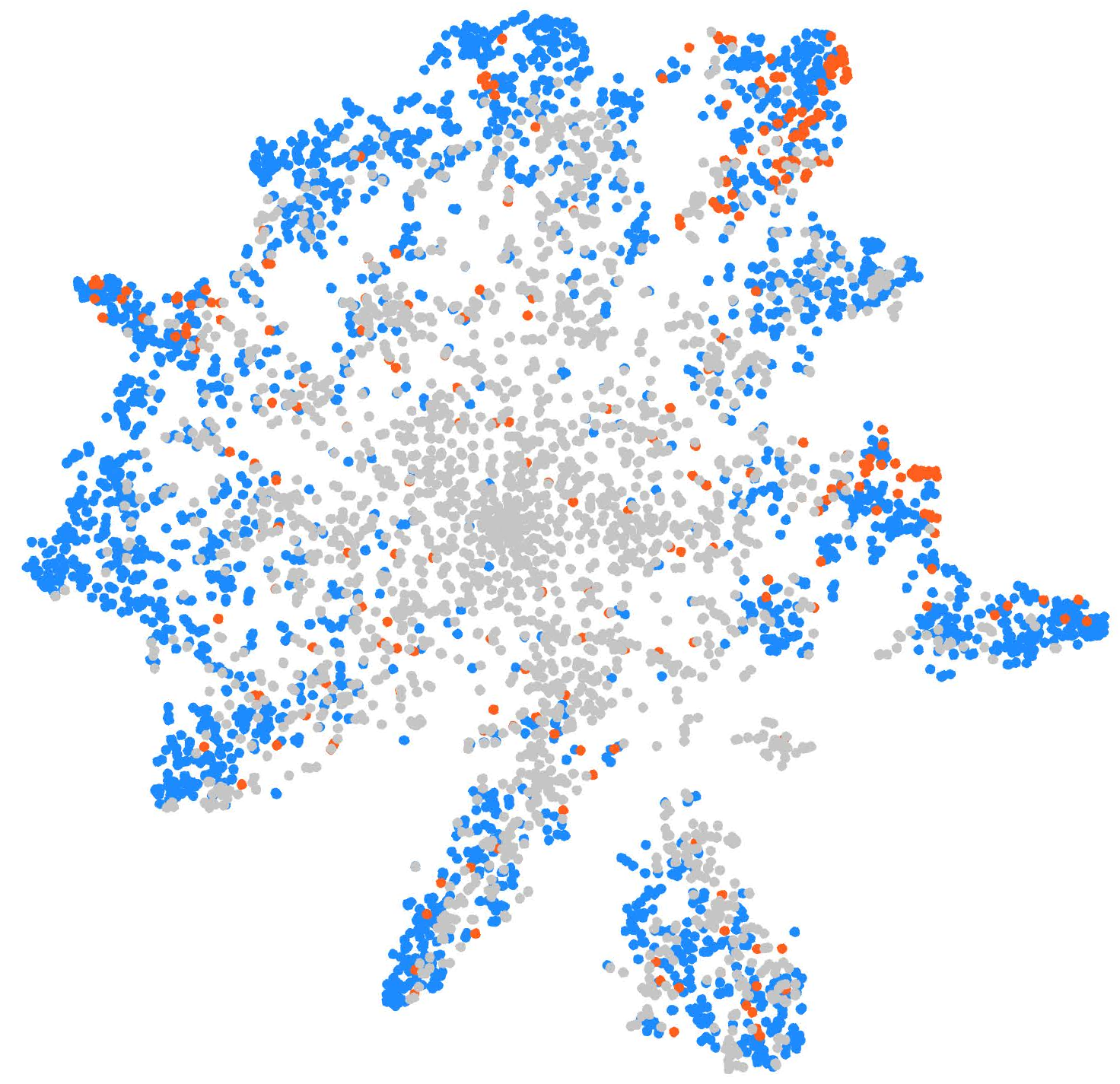}}\hspace{1ex}
    \subfloat[][STA]{\includegraphics[width=0.15\linewidth]{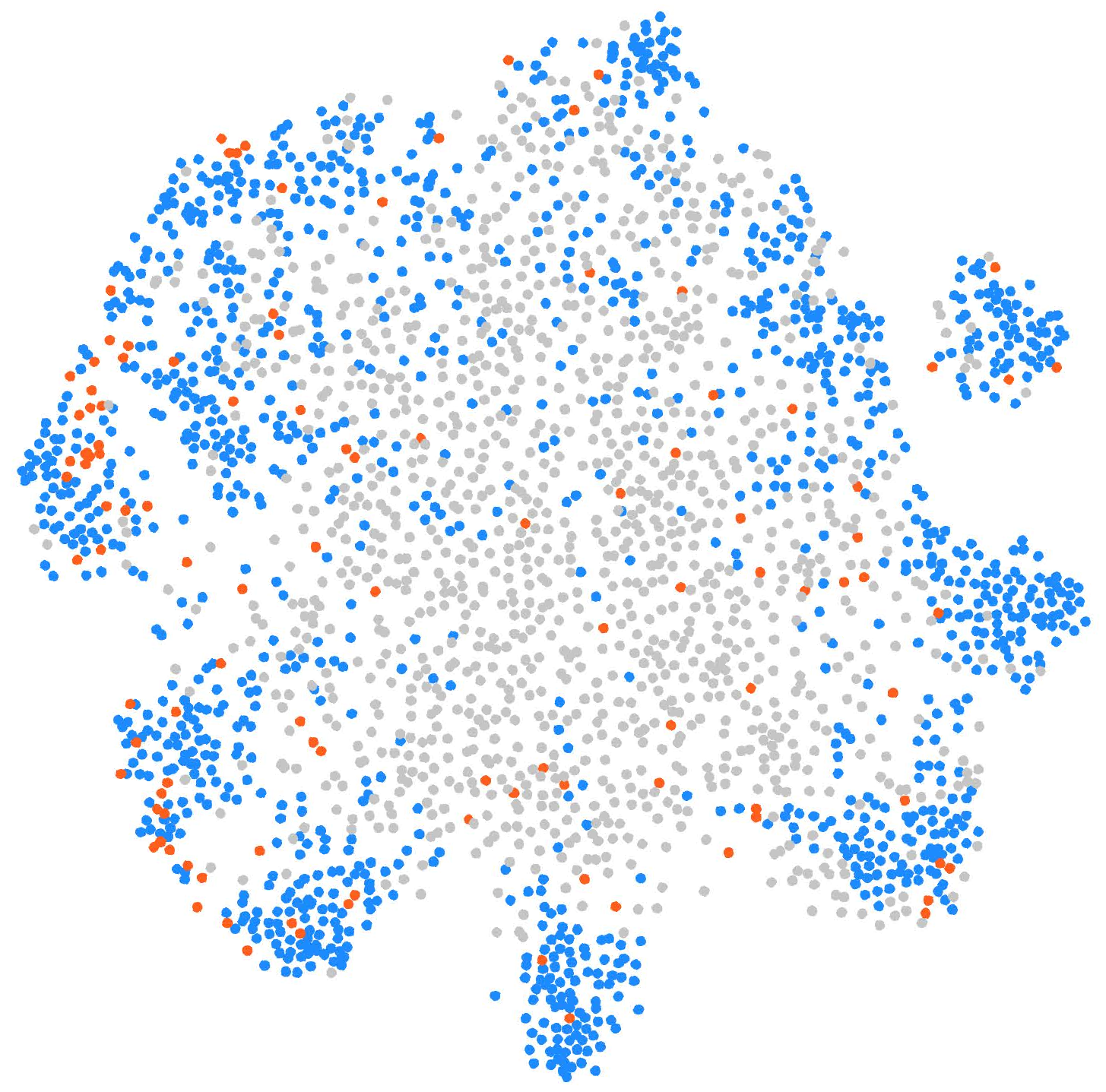}}\hspace{1ex}
    \subfloat[][PGL]{\includegraphics[width=0.15\linewidth]{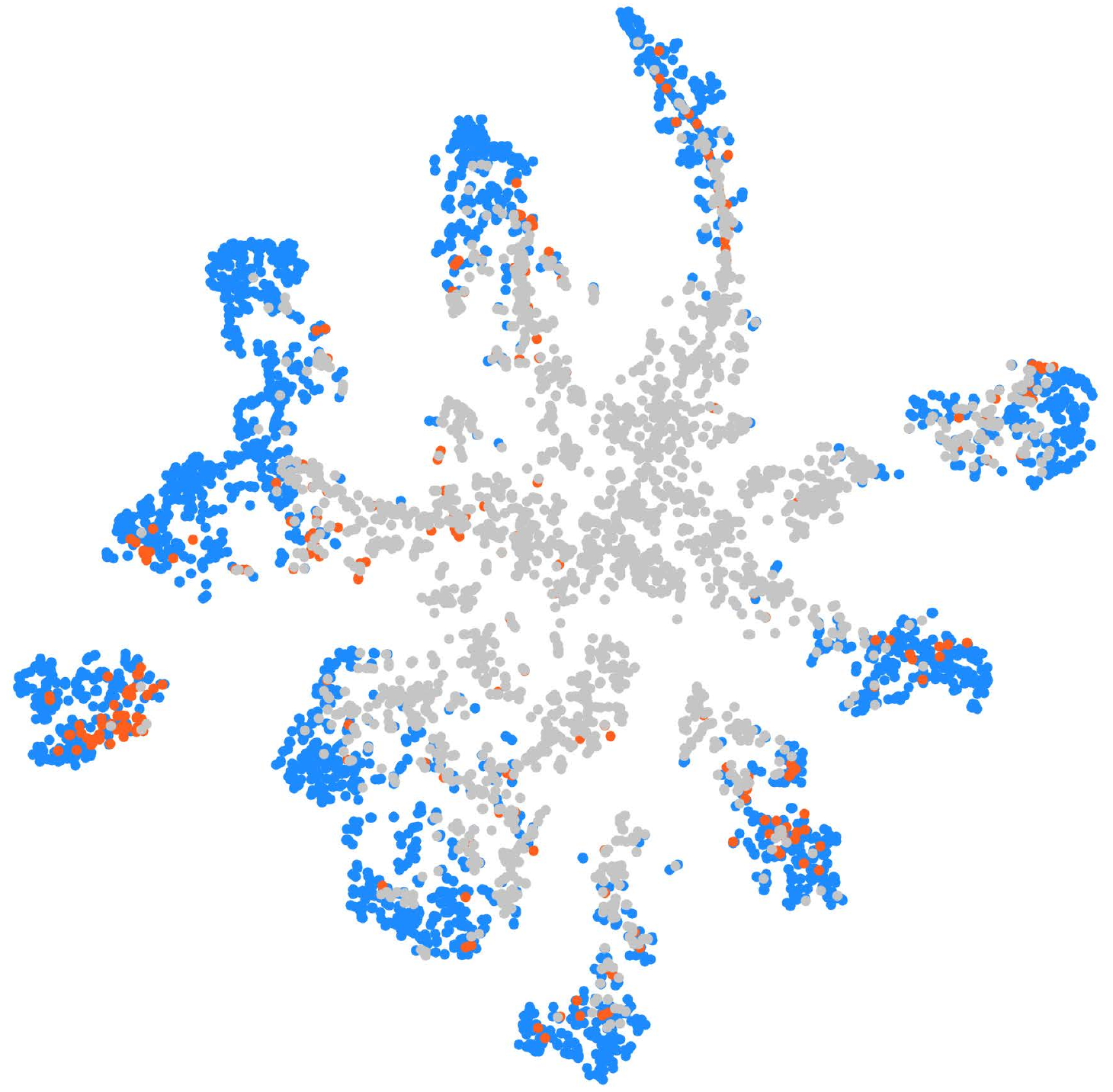}}\hspace{1ex}
    \subfloat[][SF-PGL]{\includegraphics[width=0.15\linewidth]{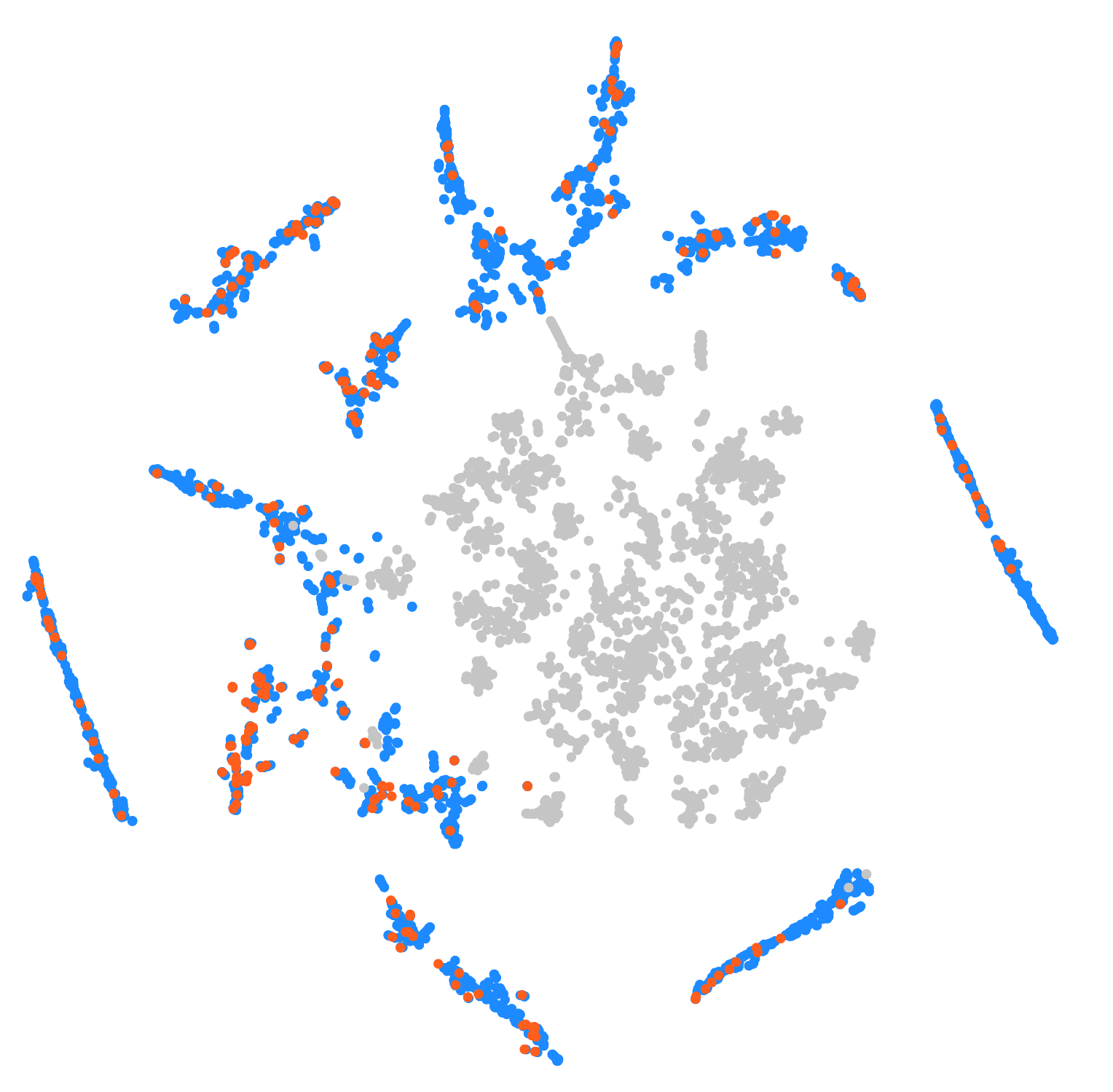}}
    \caption{The t-SNE visualization for the source and target data in the \textit{Syn2Real-O} dataset.}
    \label{fig:tsne-visda}
\end{figure*}

\begin{figure}[t]
    \centering
    \includegraphics[width=0.85\linewidth]{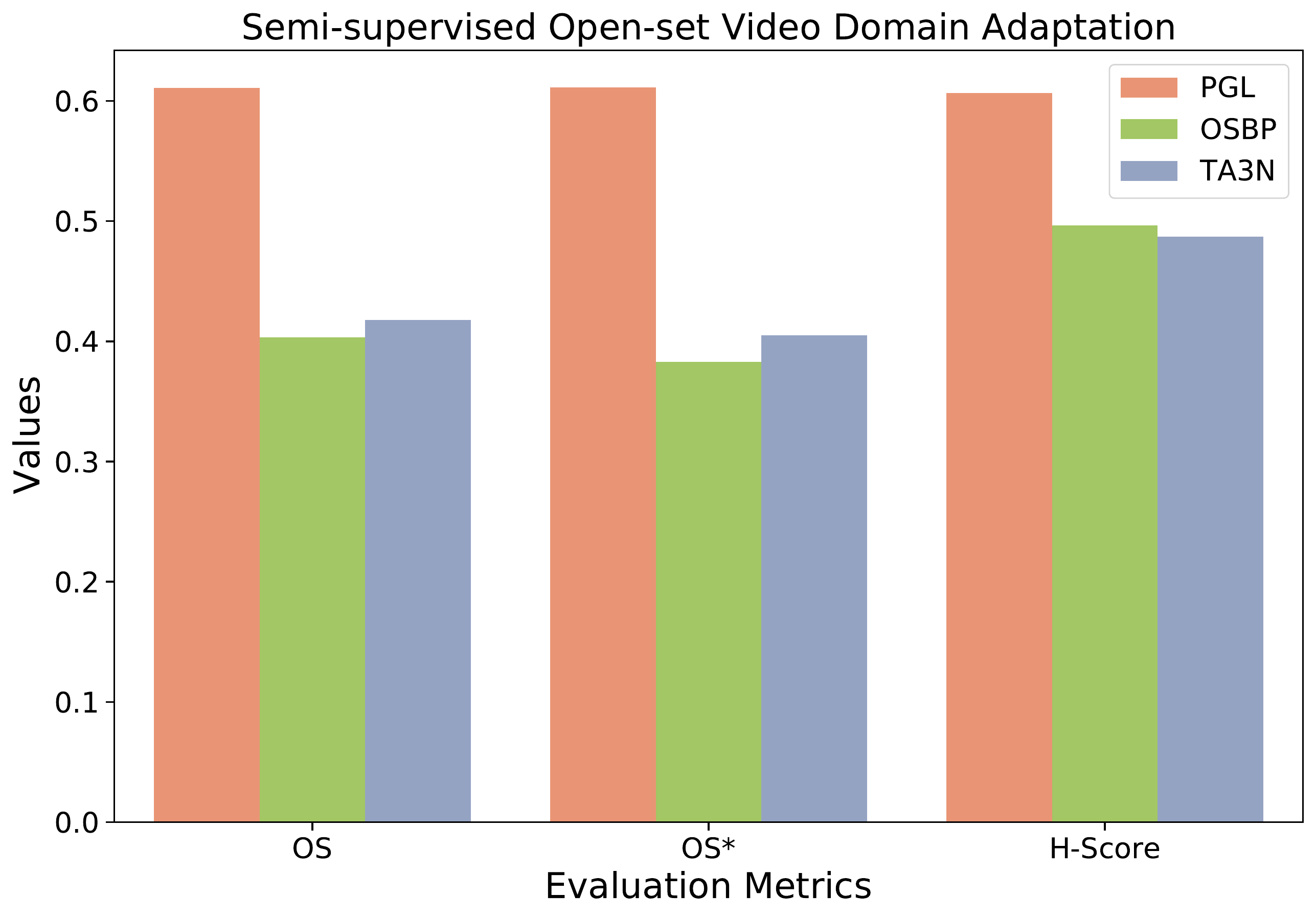}
    \caption{S-OSVDA recognition accuracies on the challenging \textit{Kinetics-Gameplay} datasets. }
    \label{fig:semiOSDA}
\end{figure}
\subsection{Results of Domain-adaptive Action Recognition}
To testify the model capacity of handling video data, we conducted open-set video domain adaptation (OSVDA) and semi-supervised (S-OSVDA) experiments on four benchmark datasets, the results of which are presented in Table \ref{tab:video} and Fig. \ref{fig:semiOSDA}. As reported in Table \ref{tab:video}, we comprehensively compare the proposed \textbf{PGL} model and the variant without adversarial learning scheme (\textbf{PGL w/o adv}) with the state-of-the-art video domain adaptation method \textbf{TA$^3$N}, open-set method \textbf{OSBP} and several closed-set domain adaptation approaches. The results show that the proposed PGL consistently outperforms the baseline and other approaches. Note that the full model gains 14.8\% and 12.7\% improvement of H-score over the variant without adversarial learning on the \textit{UCF-Olympic} and \textit{UCF-HMDB$_{full}$} datasets, respectively. We extend the unsupervised protocol to evaluate our PGL approach on a semi-supervised setting by providing $20\%$ labeled target data during training on the challenging \textit{Kinetics-Gameplay} dataset. As plotted in Figure \ref{fig:semiOSDA}, PGL obtains the best classification accuracies with regards to the OS, OS$^*$ and H scores, outperforming the state-of-the-art video domain adaptation approach TA$^3$N by 24.6\% of H score.

\subsection{Model Analysis of SF-PGL}
\textbf{t-SNE Visualization.} We conduct additional qualitative experiments of SF-PGL on the challenging \textit{Syn2Real-O} dataset with a high openness. We randomly sample 200 episodes from the dataset including 2,400 source points and 2,400 target points and visualize the distributions of the learned representations from the compared baseline models and the proposed model in Fig. \ref{fig:tsne-visda}. Fig. \ref{fig:tsne-visda}(c)-(e) shows that the open-set domain adaptation methods are more robust to disturbances from the unknowns compared with \textbf{ResNet-50} and \textbf{DANN}, as the source data (shown in blue) and the target data from the shared classes (shown in red) are aligned. A comparison between Fig. \ref{fig:tsne-visda}(d) and Fig. \ref{fig:tsne-visda}(e) reveals that the proposed graph learning and mix-up strategy have the ability to align class-specific conditional distributions across the domains, which means, the representations of the source and target data belonging to the same class are well mixed and less distinguishable. Fig. \ref{fig:tsne-visda}(f) demonstrates that the SF-PGL model can achieve tighter clusters compared with the plain PGL with a clearer separation of known and unknown samples. It is surprisingly observed from the distribution of gray points that the unknown samples are naturally formed into clusters without any labels provided, which shows the potential of further understanding the semantics of novel concepts by combining with clustering, zero-shot learning, or active learning approaches.  
\begin{figure}[t]\vspace{-1ex}
	\centering
	\subfloat[][OSBP]
	{\includegraphics[width=0.5\linewidth]{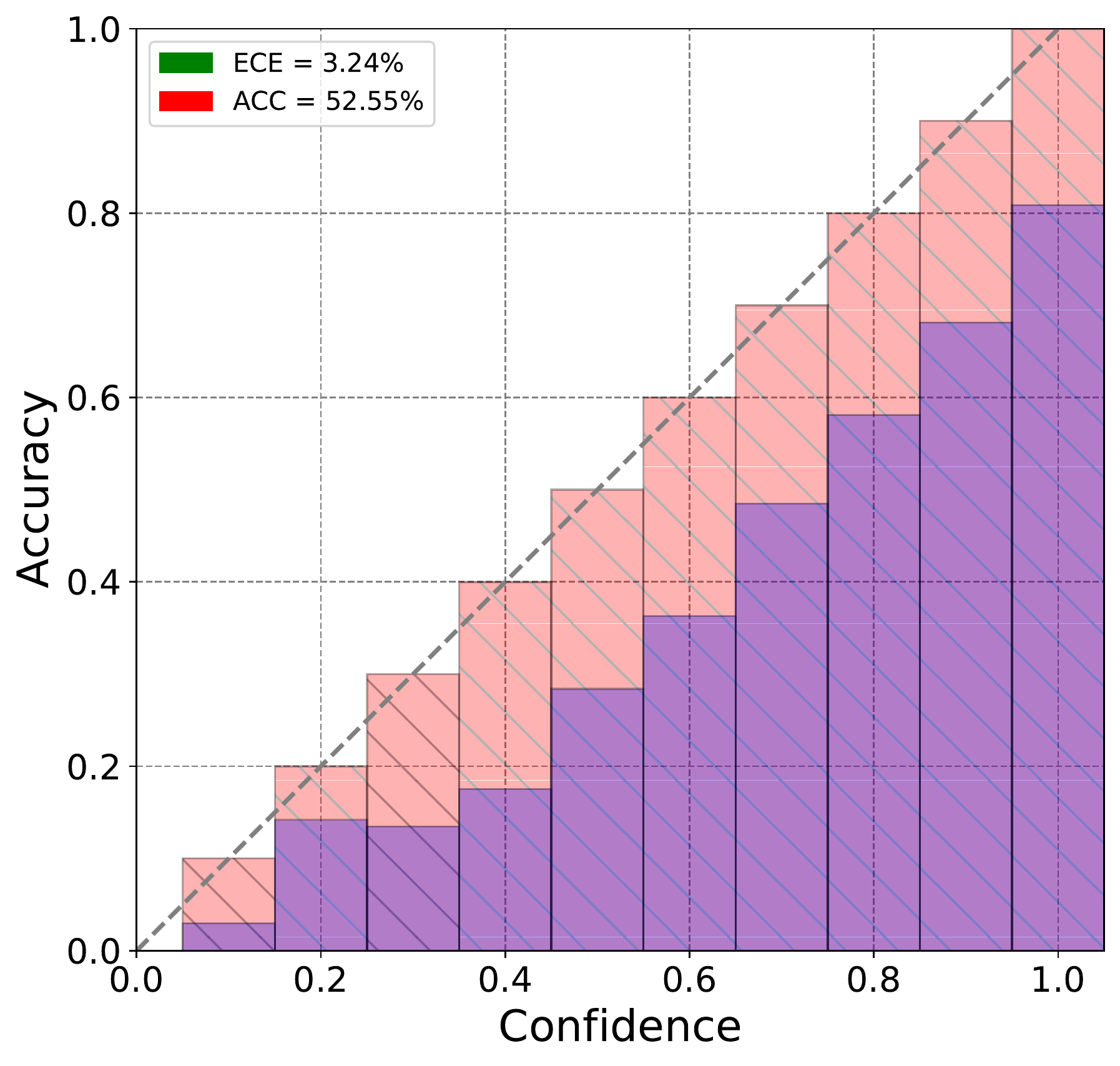}}
	\subfloat[][STA]
	{\includegraphics[width=0.5\linewidth]{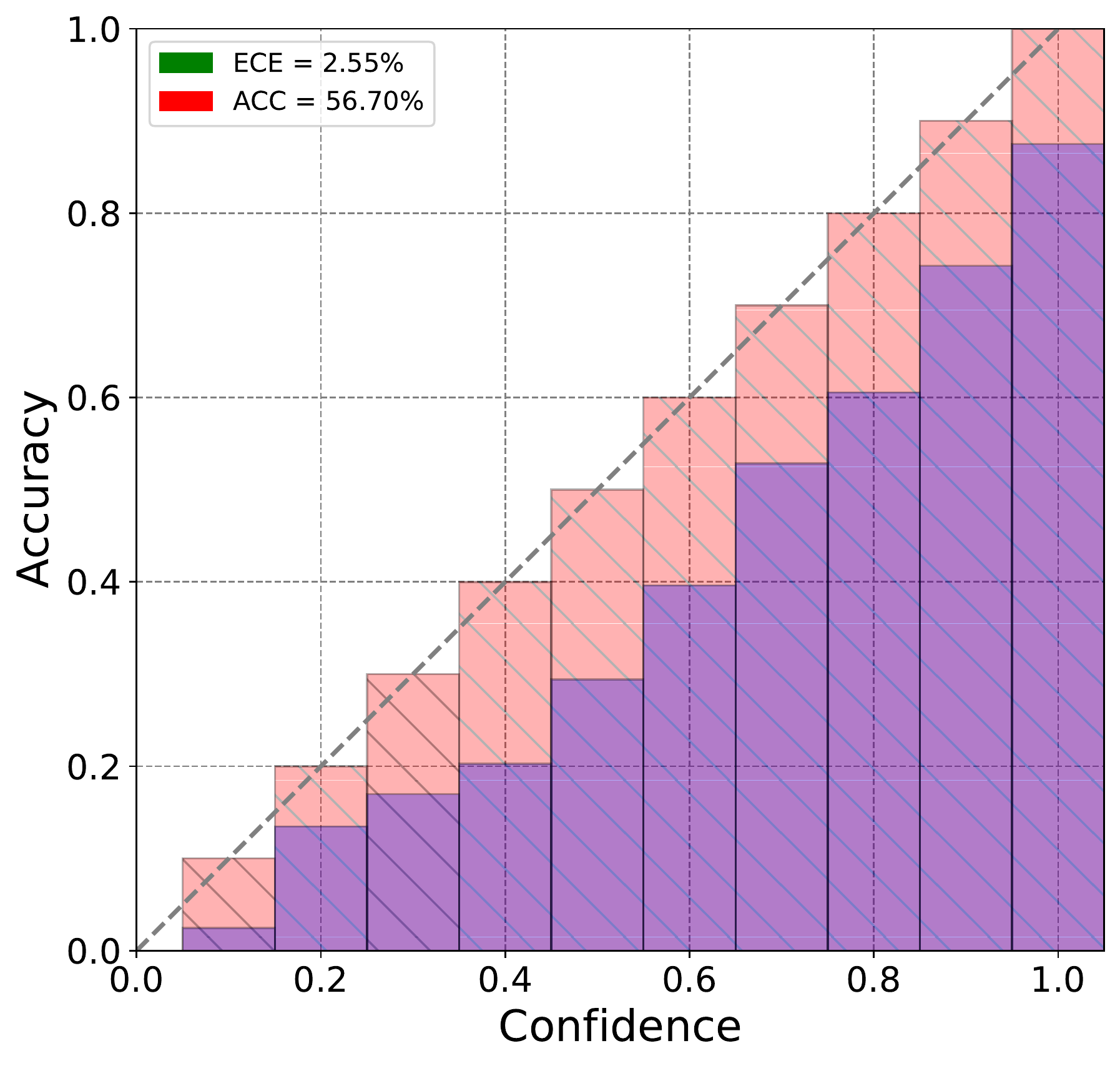}}\\
	\subfloat[][PGL]
	{\includegraphics[width=0.5\linewidth]{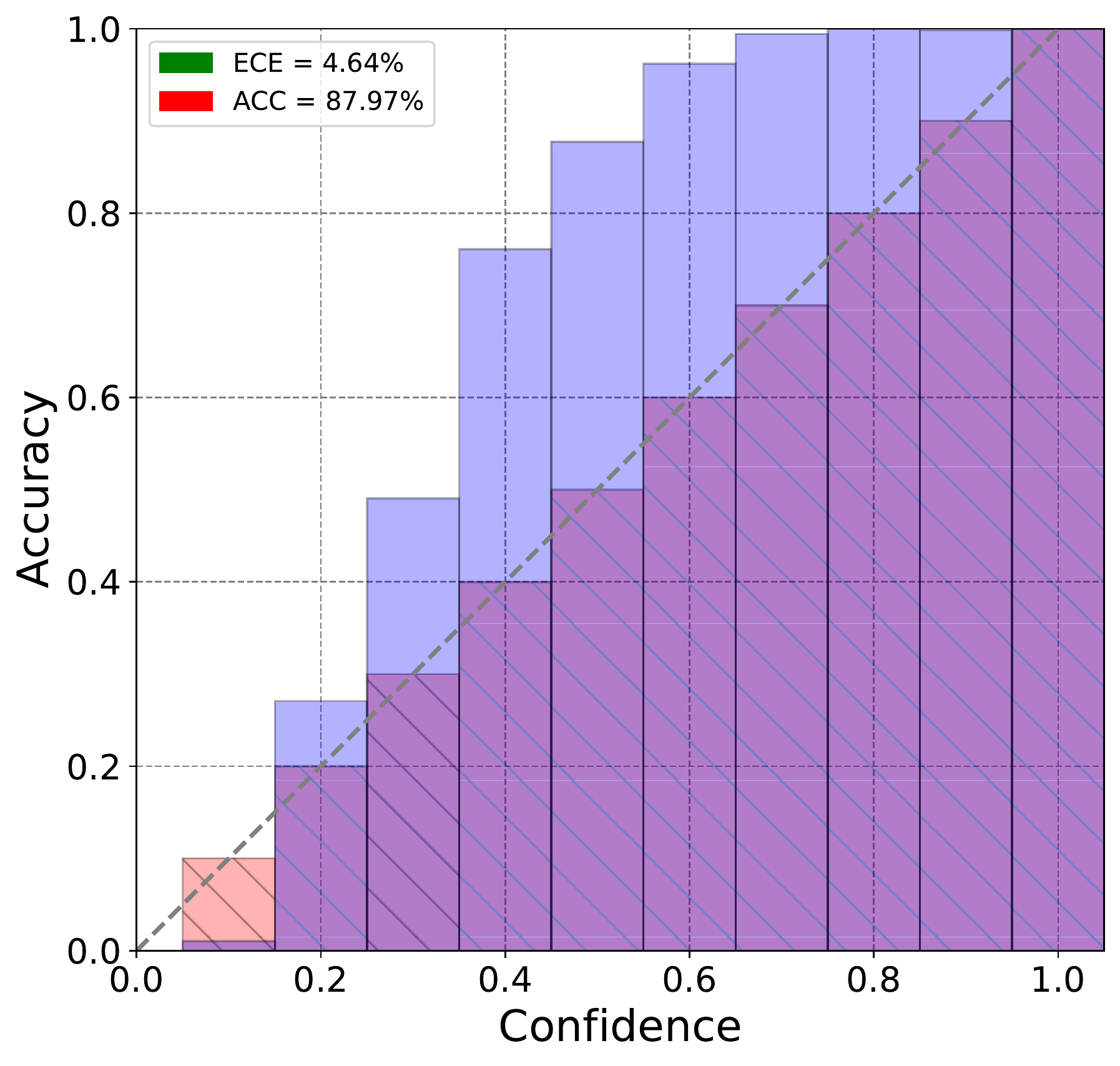}}
	\subfloat[][SF-PGL]
	{\includegraphics[width=0.5\linewidth]{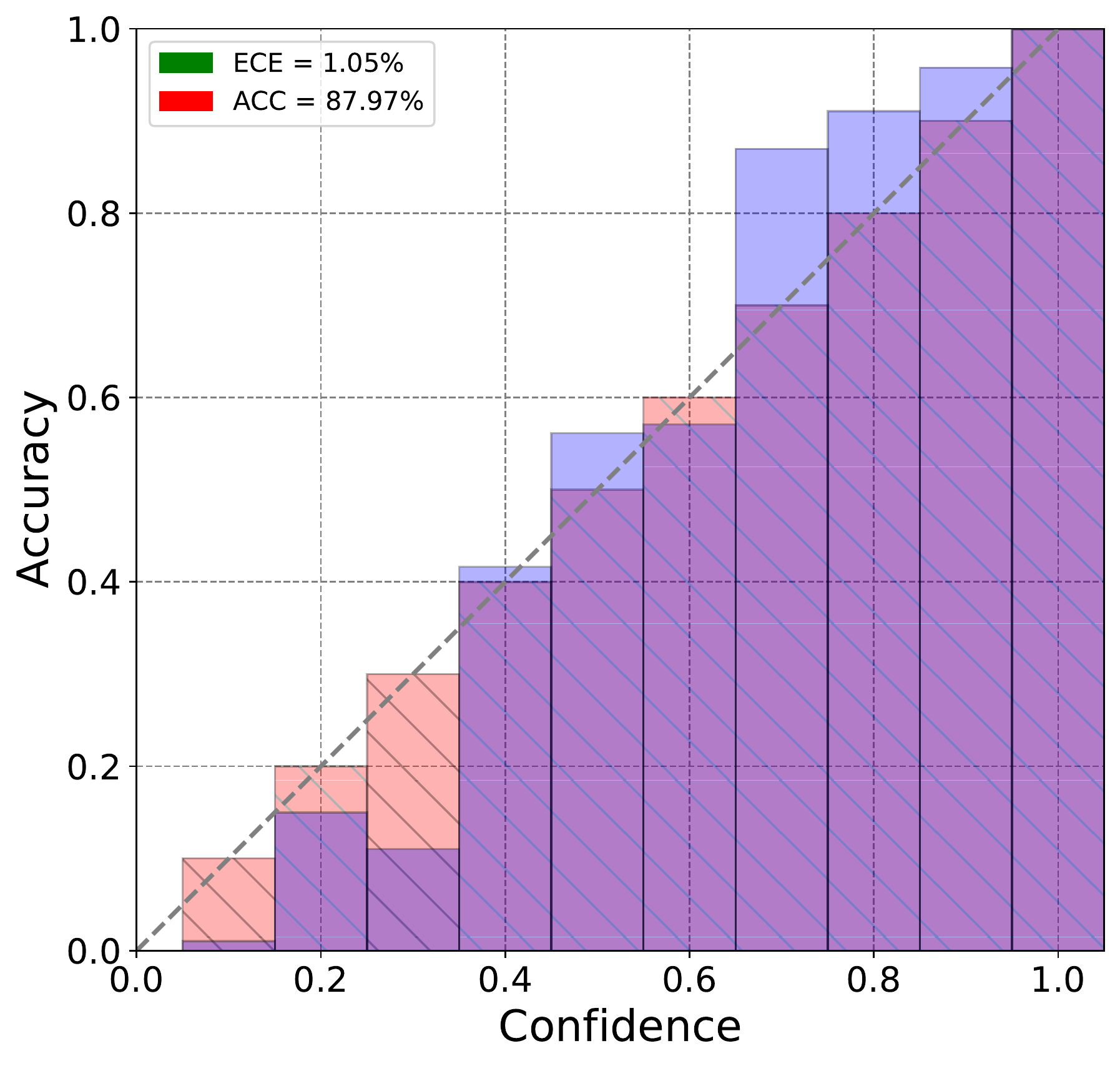}}
	\caption{Reliability diagrams for the open-set domain adaptation approaches on the \textit{Syn2Real-O} dataset.} \label{fig:calibration}
	\vspace{-1ex}
\end{figure}

\textbf{Sensitivity to Enlarging Factor $\alpha$.} To study the effect of the enlarging factor $\alpha$ on the progressive learning scheme of the SF-PGL model, we vary the value of $\alpha$ within the range $\{0.2, 0.1, 0.05\}$ on the \textit{Syn2Real-O} dataset, which leads to $\{5, 10, 20\}$ iterations for adaptation. As shown in row 5-7 and row 11-13 of Table \ref{tab:Syn2Real-O}, the model trained with a smaller value of $\alpha$ results in a steady performance increase, while along with the cost of computational time.

\textbf{Analysis of Calibration Capacity.} Figure \ref{fig:calibration} depicts the reliability diagrams for the compared open-set domain adaptation approaches and the proposed PGL and SF-PGL. Reliability diagrams \cite{DBLP:conf/icml/Niculescu-MizilC05,DBLP:conf/icml/GuoPSW17} are commonly used as a visual representation of model calibration. Based on the respective confidence scores, samples are grouped into 10 interval bins, where the expected sample accuracy and the calculated accuracy for each bin are plotted in red bars and blue bars respectively. If the model is well-calibrated, the red bars and blue bars are expected to be completely identical. Any deviation from the diagonal dotted line indicates model miscalibration. For quantitative analysis of model miscalibration, we leverage the expected calibration error (ECE) scores (Eq. \eqref{eq:ece}) as the primary empirical measurement, which captures the weighted average difference of bin’s accuracy and the confidence score. From the top row in Fig. \ref{fig:calibration}, it is observed that the compared OSBP and STA methods tend to be over-confident in the respective predictions, in a sense that the average confidence (X-axis) is substantially higher than the accuracy (Y-axis). In contrast, the PGL shown in Fig. \ref{fig:calibration}(c) is underconfident at the range (0.2, 0.9) of confidence values. This is because the high variance introduced by the pseudo labeling mechanism, and the class imbalance exacerbates the calibration problem. In Fig. \ref{fig:calibration}(d), the proposed SF-PGL improves the ECE scores from 4.64\% to 1.05\% by leveraging the balanced pseudo labeling strategy. This strategy allows the model to select the potential known data as uniform as possible, and focus on learning unconfident classes in the subsequent round.

\section{Discussion and Conclusion}
In this paper, we have proposed a generic open-set domain adaptation framework for image classification and action recognition tasks, namely progressive graph learning (PGL). By controlling the open-set risk, the proposed PGL approach addresses the domain shift in both sample- and manifold-level and the disturbance from unknown classes. To further handle a more realistic yet challenging source-free setting, a novel SF-PGL framework was proposed, which leverages a balanced pseudo-labeling regime to enable uncertainty-aware progressive learning without relying on any distribution matching or adversarial learning methods. Extensive experiments demonstrated the proposed PGL and SF-PGL framework performed consistently well on challenging source-free OSDA task and open-set action recognition task with significant domain discrepancy and conditional shifts. We further discussed a hitherto untouched aspect of OSDA model - the model calibration issue. Experimental results evidenced that the SF-PGL can alleviate the class imbalance introduced by pseudo-labeled sets so that the over-confidence and under-confidence of the OSDA model can be avoided. While the proposed SF-PGL achieved the state-of-the-art performance on large-scale OSDA benchmarks, it is still restricted from three perspectives: (1) SF-PGL cannot improve the performance for the classes if any samples from which cannot be recognized by the model pre-trained with the source data (e.g., the Truck class in Table \ref{tab:Syn2Real-O}); (2) SF-PGL is relatively $1/\alpha$ times time-consuming than OSDA methods without pseudo labeling. The observed limitations motivate future works to explore how to enhance the model generalization with only source data available, and how to minimize the steps of progressive learning without compromising performance.







\ifCLASSOPTIONcaptionsoff
  \newpage
\fi



\bibliographystyle{IEEEtran}
\bibliography{main.bib}
\end{document}